\documentclass{article}

\usepackage{PRIMEarxiv}

\usepackage{hyperref}
\usepackage{graphicx}
\usepackage{rotating}
\usepackage{times}
\usepackage{amsmath}
\usepackage{amssymb}
\usepackage{booktabs}
\usepackage{multirow}

\usepackage{subfig}
\usepackage{xfrac}
\usepackage{multirow}
\usepackage{bm}

\makeatletter
\DeclareRobustCommand*\textsubsuperscript[2]{%
	\@textsubsuperscript{\selectfont#1}{\selectfont#2}}
\def\@textsubsuperscript#1#2{%
	{\m@th\ensuremath{_{\mbox{\fontsize\sf@size\z@#1}}
			^{\mbox{\fontsize\sf@size\z@#2}}}}}
\makeatother

\usepackage{caption}
\usepackage{threeparttable}
\usepackage{nicefrac}       

\usepackage{cite}
\usepackage{lettrine}
\usepackage{epsfig}
\usepackage{tabularx,ragged2e,booktabs,caption}
\usepackage{enumitem}
\usepackage[export]{adjustbox} 
\usepackage{color}

\usepackage{array}
\newcolumntype{P}[1]{>{\centering\arraybackslash}p{#1}}

\makeatletter
\DeclareRobustCommand*\textsubsuperscript[2]{%
	\@textsubsuperscript{\selectfont#1}{\selectfont#2}}
\def\@textsubsuperscript#1#2{%
	{\m@th\ensuremath{_{\mbox{\fontsize\sf@size\z@#1}}
			^{\mbox{\fontsize\sf@size\z@#2}}}}}
\makeatother

\newcommand*{\myQED}{\hfill\ensuremath{\square}}%

\newtheorem{algorithm}{Algorithm}

\newtheorem{definition}{Definition}
\newtheorem{lemma}{Lemma}
\newtheorem{proposition}{Proposition}

\newtheorem{remark}{Remark}

\makeatletter
\def\namedlabel#1#2{\begingroup
	\def\@currentlabel{#2}%
	\label{#1}\endgroup
}
\makeatother

\makeatletter
\def\namedDeflabel#1#2{\begingroup
	\def\@currentlabel{#2}%
	\label{#1}\endgroup
}
\makeatother

\makeatletter
\def\namedLemmalabel#1#2{\begingroup
	\def\@currentlabel{#2}%
	\label{#1}\endgroup
}
\makeatother

\makeatletter
\def\namedProplabel#1#2{\begingroup
	\def\@currentlabel{#2}%
	\label{#1}\endgroup
}
\makeatother

\makeatletter
\def\namedCorolabel#1#2{\begingroup
	\def\@currentlabel{#2}%
	\label{#1}\endgroup
}
\makeatother

\makeatletter
\def\namedRemlabel#1#2{\begingroup
	\def\@currentlabel{#2}%
	\label{#1}\endgroup
}
\makeatother

\makeatletter
\def\namedAlgolabel#1#2{\begingroup
	\def\@currentlabel{#2}%
	\label{#1}\endgroup
}
\makeatother

\DeclareMathOperator*{\Ex}{\mathbb{E}}

\usepackage{breakcites}
\usepackage{etoolbox}
\makeatletter
\patchcmd{\@citex}{,}{;}{}{}
\makeatother

\graphicspath{{media/}}     

\title{A New Perspective for Understanding Generalization Gap of Deep Neural Networks Trained with Large Batch Sizes

\thanks{\textit{This work was funded by the National Research Fund (FNR), Luxembourg, under the project reference CPPP17/IS/11643091/IDform/Aouada and BRIDGES2020/IS/14755859/MEET-A/Aouada} \\ 
	\text{$^{\dagger}$ This work was done while at SnT, University of Luxembourg, Luxembourg.}} 
}

\author{
  Oyebade K. Oyedotun $^{\dagger}$ \\
  Spire Global, Luxembourg \\
  \texttt{oyebade.oyedotun@spire.com} \\
   \And
  Konstantinos Papadopoulos \\
  Post Luxembourg, Luxembourg \\
  \texttt{papad.konst@gmail.com} \\
  \AND
  Djamila Aouada \\
  Interdisciplinary Centre for Security, Reliability and Trust (SnT), \\ University of Luxembourg, Luxembourg \\
  \texttt{djamila.aouada@uni.lu} \\
}

\begin{document}
\maketitle

\begin{abstract}
Deep neural networks (DNNs) are typically optimized using various forms of mini-batch gradient descent algorithm. A major motivation for mini-batch gradient descent is that with a suitably chosen batch size, available computing resources can be optimally utilized (including parallelization) for fast model training. However, many works report the progressive loss of model generalization when the training batch size is increased beyond some limits. This is a scenario commonly referred to as generalization gap. Although several works have proposed different methods for alleviating the generalization gap problem, a unanimous account for understanding generalization gap is still lacking in the literature. This is especially important given that recent works have observed that several proposed solutions for generalization gap problem such learning rate scaling and increased training budget do not indeed resolve it. 
As such, our main exposition in this paper is to investigate and provide new perspectives for the source of generalization loss for DNNs trained with a large batch size. Our analysis suggests that large training batch size results in increased \textit{near-rank loss} of units' activation (i.e. output) tensors, which consequently impacts model optimization and generalization. Extensive experiments are performed for validation on popular DNN models such as VGG-16, residual network (ResNet-56) and LeNet-5 using CIFAR-10, CIFAR-100, Fashion-MNIST and MNIST datasets.
\end{abstract}


\section{Introduction}
Small neural network models typically contain 2 to 4 hidden layers with a moderate number of hidden units per layer~\cite{8,9,7}. Hence, small DNN models can be trained by standard gradient descent~\cite{10}, where all the available training samples are used for model updates. However, there has been a consistent increase in the difficulty of learning problems tackled over time; recent benchmarking datasets typically contain thousands or several millions of training samples, e.g, Places~\cite{11}, ImageNet~\cite{12} and UMD faces~\cite{13} datasets. Thus, there has been a consistent increase in the depth and number of DNN model parameters ever since. Subsequently, mini-batch gradient descent, which uses a specified portion of available training samples for model updates every iteration, has been favoured for training large DNN models, in view of fast training time and effective utilization of available computing resources. Therefore, fast training of large DNNs have become an important research problem. \\
One simple approach for speeding up DNN training is employing large batch sizes to fit the capacity of available Graphics Processing Unit (GPU) memory; with the parallelization of recent high-end GPU, batch sizes up to several thousands have been reported~\cite{14,15}. Unfortunately, it has also been observed that increasing the training batch size beyond some certain limits results in the degradation of the generalization of models; this performance loss with increase in batch size is commonly referred to as \textit{generalization gap}~\cite{19}. Consequently, most works~\cite{21,22} have been dedicated to proposing various approaches for reducing the generalization gap of models trained with large batch sizes; only a few works~\cite{16,20,39} have studied the cause of such performance degradation. \\
As such, in this paper, our high-level exposition is investigating the source of generalization loss observed in DNN models trained with large batch sizes. Specifically, our major contributions are as follows:
\begin{enumerate}
	\item A novel explanation and interesting insights for why DNN models trained with large batch sizes incur generalization loss based on near-rank loss of hidden units' activation tensors. This perspective on generalization gap is the first in the literature to the best of our knowledge.   	 
	\item Extensive experimental results on standard datasets (i.e. CIFAR-10, CIFAR-100, Fashion-MNIST and MNIST) and popular DNN models (i.e. VGG-16, ResNet-56 and LeNet-5) are reported for the validation of the positions given.
\end{enumerate}
The remainder of this paper is organized as follows. Related works are discussed in Section 2. In Section 3, the background and problem statement are presented. Section 4 presents our proposed analysis of the generalization gap problem. Section 5 reports the supporting experimental results. Main insights from formal and experimental results are given in Section 6. The paper is concluded in Section 7.

\section{Related work}
It is well-known that DNNs trained with large data batches have lesser generalization capacity than those trained with small batches. i.e. the \textit{generalization gap} problem. In~\cite{16}, the concept of sharp and flat minimizers is studied in relation to the generalization gap; it notes that increasing the number of training iterations does not alleviate the problem. It is further observed that DNNs trained with large data batches converge to sharp minima, while DNNs trained with small data batches converge to flat minima. Interestingly, sharp and flat minima have been shown to lead to poor and good model generalizations, respectively~\cite{17,18}. An additional explanation~\cite{16} is that noisy gradient estimation from small training data batches allow optimization to escape basins of sharp minima that the model would be stuck in for large data batches with lesser gradients stochasticity. \\
The work in~\cite{19} explores DNN optimization as high dimensional particles with random walk of random potentials; it posits diffusion rates concept for different training batch sizes. It suggests that DNNs trained with large data batches have slower diffusion rates, and thus require an exponential number of model updates to reach the flat minima regime. In~\cite{20}, it is shown that the landscape of robust optimization is flat minima and less susceptible to adversarial  attacks; this training regime is easily reached using small data batches. Further analysis shows that large training data batches resulted in models converging to solutions with a larger Hessian spectrum~\cite{20}. It is tempting to suppose that the problem of generalization gap seen in models trained with large batch sizes is related to the smaller number of parameters' updates they employ relative to models trained with small batch sizes; this direction of explanation was presented in~\cite{19,22}.  Subsequently, it may be expected that training models with large batch sizes for more epochs will indeed resolve the generalization gap problem. Unfortunately, it has been observed in~\cite{16,56} that the generalization gap persists with an unlimited training budget. The work~\cite{56} as well observed that learning rate scaling that has been proposed as a solution do not resolve generalization gap, when the batch size is very large. Subsequently, it is natural to pursue new and interesting explanations for generalization gap, which can result in the formulation of better solutions for the problem. \\
The addition of noise to computed gradients for alleviating the problem of generalization gap is seen in~\cite{58}. Therein, covariance noise was computed and added to the original gradients, and it was subsequently shown to mitigate the loss of model generalization. Another work in~\cite{59}, it was observed that gradient noise can be decomposed as the multiplication of gradient matrix and sampling noise related to how data batches are sampled. Importantly, specially computed noises were used for regularizing gradient descent and therefore model generalization. In the work~\cite{60}, for smoothing gradients computed from large-batch distributed training, local extragradient was proposed for improved model optimization for model trained with large batch sizes. The approach showed interesting results on different DNN architectures such as ResNet, LSTM and transformers. Furthermore, the work in~\cite{61}, it was reported that the ratio of batch size to learning rate negatively correlates to model generalization. As such, for good model generalization, it was advocated that the batch size should be kept as small as possible, and the learning rate keep as large a possible. It is shown in~\cite{62} that contrary to earlier claims, both SGD and second-order approximation for gradient descent, Kronecker-Factored Approximate Curvature (K-FAC), reflect generalization loss for DNNs trained with large batch sizes. 

\section{Background and Problem Statement}
This section discusses the background of the work that reflects the setting under which we investigate the generalization gap problem. Subsequently, the problem statement that specifically shows the generalization gap problem is presented. 
\subsection{Background on batch size impact for DNN}
This section discusses preliminaries for the impact of batch size on DNN generalization. Given an arbitrary dataset $D = \{(\bm{x}^n, \bm{y}^n)\}^N_{n=1} $ with input $\bm{x} \in \mathbb{R}^{h^n_0}$, target output $\bm{y} \in \mathbb{R}^c$ and sample index $n$, the pair $(\bm{x}^n, \bm{y}^n)$ is typically fed in batches as in $\bm{X} \in \mathbb{R}^{h^n_0 \times b_s}$ and $\bm{Y} \in \mathbb{R}^{c \times b_s}$ during DNN training; where $h^n_0$ and $b_s$ are the input dimension and batch size, respectively. For multilayer perceptron (MLP) models, the batch output at layer $l$ is of the form $\bm{H(X)}^l \in \mathbb{R}^{h^n_l \times b_s}$, where $h^n_l$ is the number of units in hidden layer $l$. Given the input, $\bm{H(X)}^{l-1} \in \mathbb{R}^{h^n_{l-1} \times b_s}$, to an arbitrary DNN layer $l$ parameterized by the weight $\bm{W}^l \in \mathbb{R}^{h^n_l \times h^n_{l-1}}$, we can write the transformation, $\bm{H(X)}^{l}$, learned as
\begin{equation}
	\label{eq1}
	\bm{H(X)}^{l} = \varphi(\bm{W}^l \bm{H(X)}^{l-1}),
\end{equation}
where $\varphi$ is the element-wise activation function; the bias term is omitted. Assuming the DNN has $L$ layers with output layer weight $W^L \in \mathbb{R}^{p \times h^n_L} $, the final output, $\bm{Y}$, is
\begin{equation}
	\label{eq2}
	\bm{Y} = \varphi(\bm{W}^L \varphi(\bm{W}^{L-1} \cdots \varphi(\bm{W}^1 \bm{H(X)}^0))),
\end{equation}
where $\bm{H(X)}^0$ is the input, $\bm{X}$, to the DNN.\\
\subsection{Problem statement}
Consider a DNN model denoted $\bm{M}$ with specific architecture, parameters initialization scheme and hyperparameters settings except for batch size. Let the classification test error of $\bm{M}$ be $\bm{M}_{err}^{test}$. Our main exposition in this paper is understanding why in practice, we observe the relation
\begin{equation}
	\label{eq3}
	\bm{M}_{err}^{test} \propto b_s:~~b_s \gg 1, 
\end{equation}
as seen in the work~\cite{16,19,20,56}. Increasing $b_s$ generally leads to an increase in the model error rate. How the increase in $b_s$ changes the generalization dynamics of trained DNNs has remained a challenging research question with different directions of investigation. Most of the works that have attempted to unravel the origin of generalization approached it from perspectives that decoupled model optimization and generalization performance. In contrast to earlier works, the problem of generalization gap is investigated simultaneously from both optimization and generalization perspectives.

\section{Proposed analysis of the generalization gap problem}
Herein, the proposed analysis for generalization gap in relation to the rank loss of hidden representations, information loss and optimization success are presented. However, the relevance of studying linear DNNs and basic concepts are first introduced.

\subsection{Relevance of studying linear DNNs}
The theoretical analysis of practical DNNs is generally complicated, and thus the linear activation (no non-linearities) function along with other necessary assumptions are often made for tractability. Our formal analysis in this paper assume the linear activation (i.e. linear DNNs) as well. As such, following existing literature, we first emphasize the relevance of studying linear DNNs, including their usefulness for understanding DNNs with non-linear activation functions (non-linear DNNs). In~\cite{54}, it was observed that analytically results obtained using linear DNNs conform with results from non-linear DNNs. This is expected given that the loss function of a linear multilayer DNN is non-convex with respect to the model parameters, similar to non-linear DNNs. Furthermore, the relevance of linear DNNs for analytical study is seen in the work~\cite{55}. Interestingly, we note that much stronger assumptions are common in the literature. For example, the assumption of no nonlinearities and no batch-norm can be found in~\cite{49}. In~\cite{50}, the assumptions of (i) no nonlinearities, (ii) no batch-norm, (iii) the number of units in the different layers are more than the number of units in the input or output layers, and (iv) data is whitened. 
The assumptions of (i) no batch-norm, (ii) infinite number of hidden units, and (iii) infinite number of training samples can be found in~\cite{52}. In~\cite{53}, the assumptions of (i) no nonlinearities, (ii) no batch-norm, (iii) the thinnest layer is either the input layer or the output layer, and (iv) arbitrary convex differentiable loss can be found. Nonetheless, the results from the aforementioned works have proven quite useful in practice. We show later on in our experiments that both non-linear and linear DNNs exhibit similar training characteristics for the generalization gap problem, which is not surprising.

\subsection{Preliminaries}
For analytical simplicity, we consider MLP networks that represent hidden layer units' activations with matrices. However, we show later on that the extension of analytical results to Convolutional Neural Networks (CNNs) that represents hidden layer units' activations as 4-dimensional tensors is straightforward.\\
The rank of hidden layer units' activations, $\bm{H(X)}^l \in \mathbb{R}^{h^n_l \times b_s}: h^n_l \geq b_s$, describes the maximum number of linearly independent columns of $\bm{H(X)}^l$. The rank of $\bm{H(X)}^l$ can be obtained as its number of non-zero singular values, $S^l_{nz}$, via singular value decomposition (SVD). Specifically, a matrix with full rank means that all its singular values are non-zero. Conversely, the number of zero singular values, $S^l_z$, shows the degree of rank loss. 
\begin{definition}
	\namedDeflabel{def 1}{Definition }
	\textit{Near-rank loss} in this paper is taken to mean a scenario where the singular values denoted by $\sigma$ are very small (i.e. $\sigma \ll 1$). 
\end{definition}
Our main analytical observations for the problem of generalization gap are summarized as follows in Section 4.3. 
\subsection{Units' activation near-rank loss and optimization}
In this section, the limits of the singular values of random matrices are related to their dimensions. For stating the first proposition, we consider that the hidden layer representations, $\bm{H(X)^l} \in \mathbb{R}^{h^n_l \times b_s}$, are samples of some unknown distribution, $p(\bm{H(X)}^l) $, and then characterize the behaviour of the maximum and minimum singular vlaues for $b_s \longrightarrow \infty$. In the second proposition, we assume that the enteries in $\bm{H(X)^l} \in \mathbb{R}^{h^n_l \times b_s}$ follow a Gaussian distribution. Subsequently, we characterize the limits of the singular values of $\bm{H(X)^l}$ for $b_s \in \mathbf{R}$.

\begin{proposition}
	\namedProplabel{prop 1}{Proposition 1}
	(The asymptotic behaviour of the singular values of matrix with increase in dimension): For a matrix $\bm{A} \in \mathbb{R}^{m \times n}: m \geq n$,~from Marchenko-Pastur law,~the singular values, $\sigma$,~concentrate in the range $[\sigma_{min}(\bm{A}) \sim \sqrt{m}-\sqrt{n}, \sigma_{max}(\bm{A}) \sim \sqrt{m}+\sqrt{n}]$ as $ m, n \longrightarrow \infty $; where $\sigma_{max}(\bm{A})$ and $\sigma_{min}(\bm{A})$ are the maximum and minimum singular values of $\bm{A}$, respectively. 
\end{proposition} 
\noindent \textit{Proof.} See Rudelson-Vershynin~\cite{34} for proof. \myQED

\begin{remark}
	\namedRemlabel{rem 1}{Remark 1}
	In fact,~\cite{34} notes that~\ref{prop 1} holds for general distributions. As such, we can conclude from Proposition 1 that $\bm{H(X)^l} \in \mathbb{R}^{h^n_l \times b_s}: b_s \longrightarrow \infty $ results in small and large distribution ranges, which are admissible for $\sigma_{min}(\bm{H(X)}^l)$ and $\sigma_{max}(\bm{H(X)}^l)$, respectively. Accordingly, as $b_s$ increases, we have the following scenarios (i) higher probability for $\bm{H(X)}^l$ to have a small $\sigma_{min}(\bm{H(X)}^l)$; and (ii) higher probability for $\bm{H(X)}^l$ to have a larger condition number. 
\end{remark}

\begin{proposition}
	\namedProplabel{prop 2}{Proposition 2}
	(The non-asymptotic behaviour of the singular values of matrix with increase in dimension): For a Gaussian random matrix $\bm{A} \in \mathbb{R}^{m \times n}: m \geq n$, the expected minimum and maximum singular values are given as
	\begin{equation}
		\label{eq3b}
		\sqrt{m}-\sqrt{n} \leq \Ex \sigma_{min} (\bm{A}) \leq \Ex \sigma_{max} (\bm{A}) \leq \sqrt{m}+\sqrt{n}
	\end{equation}	
	
\end{proposition} 
\noindent \textit{Proof.} See Theorem 2.6 in Rudelson-Vershynin~\cite{34}. \myQED \\
In Section A1.1 and Section A1.2 of the supplementary material, we empirically study the distributions of singular values and expected values of the minimum singular values for random matrices, respectively, where the entries are drawn from popular distributions including the Gaussian, uniform and lognormal. Subsequently, for the aforementioned distributions, in Section A1.2 of the supplementary material, we show that for a fixed $m$, $\Ex \sigma_{min} (\bm{A}) \longrightarrow 0$, as $n$ becomes large. This observation is akin to that in~Eqn.\eqref{eq3b}, so that it is applicable for other popular distributions. 
\begin{remark}
	\namedRemlabel{rem 2}{Remark 2}
	Given $\bm{H(X)^l_1} \in \mathbb{R}^{h^n_l \times {b_s}_1}: h^n_l > {b_s}_1$ and $\bm{H(X)^l_2} \in \mathbb{R}^{h^n_l \times {b_s}_2}: h^n_l > {b_s}_2$ with ${b_s}_2 > {b_s}_1$, then $\Ex \sigma_{min} (\bm{H(X)^l_1}) > \Ex \sigma_{min} (\bm{H(X)^l_2})$ using~\ref{prop 2}. Importantly, given that $h^n_l$ is fixed as it is typical for DNNs,~\ref{prop 2} shows that for $\bm{H(X)^l} \in \mathbb{R}^{h^n_l \times b_s}: b_s \gg 1$, we have $\Ex \sigma_{min} (\bm{H(X)^l_2}) \ll 1$; that is, near-rank loss. 
\end{remark}
~\ref{prop 1} and~\ref{prop 2} show the asymptotic and non-asymptotic behaviours of the extreme singular values of matrices with respect to their dimensions, respectively. In both cases, it is observed that the characteristic of $\sigma_{min} (\bm{H(X)^l})$ is similar. That is, as $b_s$ increases, $\sigma_{min} (\bm{H(X)^l})$ decreases.

\begin{proposition}
	\namedProplabel{prop 3}{Proposition 3}
	Assuming an $L$-layer linear DNN is parameterized by $\bm{\theta} = \{\bm{W}\}_{l=1}^L$ (where $\bm{\theta}$ is invertible), and its input data is $\bm{X} \in \mathbb{R}^{h^n_0 \times b_s}$ so that the output at hidden layer $l$ is $\bm{H(X)}^l \in \mathbb{R}^{h^n_l \times b_s}$. A small $\Delta \bm{H(X)}^l$ translates to the small solution change, $\lVert \Delta \bm{\theta} \rVert$, so that $\lVert \Delta \bm{\theta} \rVert / \lVert \bm{\theta} \rVert$ is given as
	\begin{equation}
		\label{eq4}
		\frac{\lVert \Delta \bm{\theta} \rVert}{\lVert \bm{\theta} \rVert} \leq \Sigma_{i=1}^r \frac{1}{\sigma^l_i}  \lVert \Delta \bm{H(X)}^l \rVert \lVert \bm{v}^l_i {\bm{u}^l_i}^T \rVert  ~~:0 \leq l \leq L,
	\end{equation}
	where $\bm{H(X)}^0 = \bm{X}$ for $l=0$, ${\bm{u}^l_i}^T \in \mathbb{R}^{1 \times h^n_l}$ and $\bm{v}^l_i \in \mathbb{R}^{b_s \times 1}$ are the left and right singular vectors of $\bm{H(X)}^l$, respectively; $\sigma^l_i$ is the singular value for $\bm{H(X)}^l$, and $T$ denotes transpose. 
\end{proposition} 
\noindent \textit{Proof.} Considering the simple problem $\bm{Y}=\bm{\theta X}$, where the objective is to estimate the solution $\bm{\theta} \in \mathbb{R}^{h^n_0 \times h^n_0}$ with $\bm{X} \in \mathbb{R}^{h^n_0 \times b_s}$ and $\bm{Y} \in \mathbb{R}^{h^n_0 \times b_s}$; where $\bm{X}$ is of rank $r$. Let a small change $\Delta \bm{X}$ result in the small solution change $\Delta \bm{\theta}$. Thus, we can write
\begin{equation}
	\label{eq5}
	Y = (\bm{\theta} + \Delta \bm{\theta}) (\bm{X} + \Delta \bm{X}),
\end{equation}
\begin{equation}
	\label{eq6}
	Y = (\bm{\theta}\bm{X} + \bm{\theta} \Delta \bm{X} + \Delta \bm{\theta} \bm{X} + \Delta \bm{\theta} \Delta \bm{X}).
\end{equation}
Noting that $\bm{Y}=\bm{\theta X}$ and $\Delta \bm{\theta} \Delta \bm{X} \approx 0$ in~Eqn.\eqref{eq6}, and taking $\bm{X}^{\dag}$ as the pseudoinverse of $\bm{X}$, we obtain
\begin{equation}
	\label{eq7}
	\Delta \bm{\theta} \bm{X} = - \bm{\theta} \Delta \bm{X},
\end{equation}
\begin{equation}
	\label{eq8}
	\bm{\theta}^{-1} \Delta \bm{\theta} = - \Delta \bm{X} \bm{X}^{\dag}.
\end{equation}
Employing Cauchy-Schwarz inequality for~Eqn.\eqref{eq8} gives
\begin{equation}
	\label{eq8b}
	\frac{\lVert \Delta \bm{\theta} \rVert}{\lVert \bm{\theta} \rVert} \leq \lVert \Delta \bm{X} \rVert \lVert \bm{X}^{\dag} \rVert.
\end{equation}

Finally, using $\bm{X}^{\dag} = \Sigma_{i=1}^r \dfrac{1}{\sigma_i} \bm{v}_i \bm{u}_i^T$ (since $\bm{X} = \Sigma_{i=1}^r \sigma_i  \bm{u}_i \bm{v}_i^T$) from SVD in~Eqn.\eqref{eq8b} completes the proof. \myQED 

\begin{remark}
	\namedRemlabel{rem 3}{Remark 3}
	From~\ref{prop 3}, it is seen that $\sigma^l_i \ll 1$ (i.e. near-rank loss) magnifies small $\Delta \bm{H(X)}^l$ so that the bound on $\lVert \Delta \bm{\theta} \rVert / \lVert \bm{\theta} \rVert$ increases unreasonably. Importantly, it is seen from (4) that the impact of $\sigma^l_i$ on $\lVert \Delta \bm{\theta} \rVert / \lVert \bm{\theta} \rVert$ is additive, and can be very pronounced based on the number of $\sigma^l_i \ll 1$. i.e. $S^l_z$. The implication of this during the training of a $L$-layer DNN parameterized by $\bm{\theta}$ is that $\lVert \Delta \bm{\theta} \rVert / \lVert \bm{\theta} \rVert$ fluctuates drastically for small $\Delta \bm{H(X)}^l: 0 \leq l \leq L$ so that optimization may not reach one of the best minima in the parameter space, $\bm{\theta}$; \textit{that is, many $S_z$ (i.e. $\sigma^l_i \ll 1$) results in worse optimization}. Ideally, we would like $\sigma^l_i \approx 1$, so that a small change $\Delta \bm{H(X)}^l$ results in a reasonably small relative change in the solution, $\lVert \Delta \bm{\theta} \rVert / \lVert \bm{\theta} \rVert$. We show via experiments that the degree of near-rank loss of hidden representations, measured by $S_z$, is correlated with batch size and achievable training loss.
\end{remark}

\begin {table} [t!]	

\begin{algorithm}
	\namedAlgolabel{algo 1}{Algorithm 1}
	\centering
	\fontsize{9}{9.5}\selectfont
	\begin{tabular}{lllll}
		\toprule
		Definition: Batch size in relation to the near-rank loss of the hidden \\ representations of DNNs. \\
		Input: $\bm{X} \in \mathbb{R}^{h^n_0 \times N}$: $N$ is the size of training set. \\
		Output: $S^l_z$, the number of singular values below $t_{th}$ in layer $l$. \\
		\midrule
		1) Choose, $b_s: 1 \leq b_s \leq N$, and $t_{th}$ (threshold for singular values) \\
		2) Initilize the counter for singular values, $S^{total}_z= 0$ \\
		3) Feed a batch of input data, $\bm{X} \in \mathbb{R}^{h^n_0 \times b_s}$, into the DNN \\
		For $l$ = 1: $L$ \\
		~~~~i) Get the output at layer $l$, $\bm{H(X)}^l \in \mathbb{R}^{h^n_l \times b_s}$ \\
		~~~~ii) Compute the minimum singular values for $\bm{H(X)}^l$  \\ 
		~~~~iii) Determine the number of singular values below $t_{th}$, $S^l_z$ \\
		~~~~iv) Accumulate $S^l_z$ for layer $l$, as in $S^{total}_z = S^{total}_z + S^l_z$ \\
		4) Return $S^{total}_z$ that summarizes the degree of near-rank \\ loss of the DNN \\
		\bottomrule
	\end{tabular}
\end{algorithm}
\vskip .1cm
\label{tab:title}	
\end {table}

\subsection{Computing singular values for hidden representations}
For multilayer perceptron (MLP) networks, where the hidden representations, $\bm{H(X)}^l \in \mathbb{R}^{h^n_l \times b_s}$, are matrices, the conventional Singular Value Decomposition (SVD) can be used to determine the singular values of the hidden representations in the different layers of the model. \\
Interestingly, most computer vision tasks rely on CNNs, where the hidden representations are tensors that can be expressed in the form $\bm{H(X)}^l \in \mathbb{R}^{d^h_l \times d^w_l \times n^c_l \times b_s}$; where $d^h_l$, $d^w_l$ and $n^c_l$ are the channel height, channel width and number of channels in layer $l$, respectively \footnote{Setting $d^h_l =1$ and $d^w_l=1$ transforms the CNN into the MLP model \label{1}}. In order to determine the singular values of the hidden representations of CNNs, we employ the Higher Order SVD (HOSVD)\footnote{Otherwise known in the literature as multilinear SVD \label{2}}\cite{29,45}, which generalizes SVD for matrices to $n$-dimensional tensors. Specifically, given the CNN hidden representation in layer $l$, $\bm{H(X)}^l$, that is a tensor, HOSVD is used for determining the singular values. For determining the degree of near-rank loss, we count the number of singular values, $S^l_z$, that below a specified threshold, $t_{th}$. In the experiments, we consider $t_{th} = 10^{-4}$ and $t_{th} = 10^{-5}$. \\
\ref{algo 1} summarizes the evaluation of the degree of near-rank loss of the DNNs used for experiments in this paper.

\subsection{Batch size in relation to the dimension of the layer weights and hidden representations}
In Section 4.4, it is seen that the dimensions of the hidden representations in both MLPs and CNNs depend on the chosen batch size. In this section, we emphasize that the dimensions of the weights in both MLPs and CNNs are unaffected by the batch size hyperparameter. This clarification is important for the interpretation of the experimental results reported in Section 5. \\
Let the weight in layer $l$ of a MLP network be $\bm{W}^l_{MLP} \in \mathbb{R}^{h^n_l \times h^n_{l-1}}$, and the weight in the layer $l$ of a CNN be $\bm{W}^l_{CNN} \in \mathbb{R}^{d^h_l \times d^w_l \times n^c_{l-1} \times n^c_l}$, where $n^c_{l-1}$ is the number of channels in layer $l-1$. \textit{It is observed from $\bm{W}^l_{MLP}$ and $\bm{W}^l_{CNN}$ that the chosen batch size, $b_s$, does not affect the dimension of the parameter space irrespective of the model architecture; instead, the dimension of the hidden (i.e. latent space) representation is affected; see Section 4.4}.

\section{Experiments}
In this section, the different datasets, DNN architectures, training settings and experimental results that can be used for understanding the generalization problem are presented. 
\subsection{Datasets}
CIFAR-10~\cite{25}, CIFAR-100~\cite{25}, Fashion-MNIST~\cite{26} and MNIST~\cite{27} datasets are used for experiments. CIFAR-10 and CIFAR-100 both have 50,000 training and testing images; CIFAR-10 and CIFAR-100 have 10 and 100 classes, respectively. Fashion-MNIST and MNIST datasets both have 60,000 and 10,000 training and testing images, respectively. In addition, both datasets have 10 different classes. We observe that similar datasets have been used in~\cite{16,20} for studying the generalization gap problem.

\subsection{DNN architectures and hyperparameters settings}
Four benchmarking datasets with three popular DNN architectures are used to generalize our findings. We perform experiments using VGG-16~\cite{28} on CIFAR-10 and CIFAR-100 datasets, ResNet-56~\cite{28} on CIFAR-10 and CIFAR-100 datasets and LeNet-5~\cite{28} on Fashion-MNIST and MNIST datasets. VGG-16 and ResNet-56 are trained on the CIFAR datasets, since they are harder datasets as compared to Fashion-MNIST and MNIST datasets. \\
We note that the claim that \textquotedblleft simply scaling the initial base learning rates in relation to the chosen batch size generally resolves the generalization problem\textquotedblright~has been refuted in~\cite{38,39}. In fact, the work in~\cite{38} further argues that there is no standard rule for hyperparameters setting that works for all datasets and models in mitigating the generalization gap problem. Equipped with these observations, our experiments are thus tailored toward studying the rank loss of hidden representations of models trained with small and large batch sizes under similar hyperparameter settings.\\
All models are trained using mini-batch gradient descent with a momentum rate of 0.9. The initial learning rate of all DNNs are set to 0.1, and are annealed to $10^{-4}$ during training. All the experiments use batch normalization~\cite{37}. A weight decay penalty of $1 \times 10^{-4}$ is used for regularizing models trained on the CIFAR datasets; a weight decay penalty of $5 \times 10^{-4}$ is used for Fashion-MNIST and MNIST datasets. Models for the CIFAR datasets are trained for 200 epochs, while models for Fashion-MNIST and MNIST datasets are trained for 60 epochs. For studying how batch size impacts optimization and generalization, we experiment with batch sizes of 64, 128, 512, 1024 and 2048 for the CIFAR datasets; and batch sizes of 128, 512, 1024, 2048, 4096 and 8192 for Fashion-MNIST and MNIST datasets. Note that batch sizes smaller than 64 are not considered for study, since they are of little practical interest (i.e. inefficient) for training modern DNNs that are typically very large and deep. For determining the near-rank loss of hidden layer units' activations tensors, a small threshold value of $t_{th} = 10^{-4}$ for singular values is used in the main experiments. Specifically, we \textit{compute the number of singular values, $S_z^{total} = \Sigma^L_{l=1} S^l_z $, that are below the threshold value for every layer in a model, and report the sum on a log-scale; see~\ref{algo 1}}.

\begin {table} [t!]	
\centering
\fontsize{9}{9}\selectfont
\begin{tabular}{|c|c|c|c|c|}
	\toprule
	Batch size & Train error & Test error & Train loss & Test loss \\	
	\midrule
	64 & 0.01\% & 6.26\% & \textbf{0.1675} & 0.4914\\
	128 & 0.02\% & 6.57\% & \textbf{0.1694} & 0.4964\\
	512 & 0.02\% & 6.73\% & \textbf{0.2198} & 0.5597\\
	1024 & 0.03\% & 7.81\% & \textbf{0.3495} & 0.7221\\
	2048 & 0.05\% & 9.12\% & \textbf{0.5210} & 0.9157\\
	\bottomrule
\end{tabular}
\vskip .2cm
\caption {VGG-16 results on CIFAR-10 dataset with different batch sizes} 
\label{tab:title}	
\end {table}

\begin {table} [t!]	
\centering
\fontsize{9}{9}\selectfont
\begin{tabular}{|c|c|c|c|c|}
	\toprule
	Batch size & Train error & Test error & Train loss & Test loss \\	
	\midrule
	64 & 0.02\% & 26.97\% & \textbf{0.4635} & 1.9880\\
	128 & 0.04\% & 27.54\% & \textbf{0.4763} & 1.9901\\
	512 & 0.05\% & 30.12\% & \textbf{0.5737} & 2.2200\\
	1024 & 0.05\% & 32.05\% & \textbf{0.6086} & 2.3345\\
	2048 & 0.07\% & 35.29\% & \textbf{0.6743} & 2.4935\\
	\bottomrule
\end{tabular}
\vskip .2cm
\caption {VGG-16 results on CIFAR-100 dataset with different batch sizes} 
\label{tab:title}	
\end {table}

\begin {table} [t!]	
\centering
\fontsize{9}{9}\selectfont
\begin{tabular}{|c|c|c|c|c|}
	\toprule
	Batch size & Train error & Test error & Train loss & Test loss \\	
	\midrule
	64 & 0.05\% & 6.02\% & \textbf{0.1173} & 0.4195\\
	128 & 0.13\% & 6.63\% & \textbf{0.1426} & 0.4586\\
	512 & 0.20\% & 8.10\% & \textbf{0.1621} & 0.5351\\
	1024 & 0.24\% & 9.27\% & \textbf{0.1959} & 0.6275\\
	2048 & 0.66\% & 10.68\% & \textbf{0.2677} & 0.7070\\ 
	\bottomrule
\end{tabular}
\vskip .2cm
\caption {ResNet-56 results on CIFAR-10 dataset with different batch sizes} 
\label{tab:title}	
\end {table}

\begin {table} [t!]	
\centering
\fontsize{9}{9}\selectfont
\begin{tabular}{|c|c|c|c|c|}
	\toprule
	Batch size & Train error & Test error & Train loss & Test loss \\	
	\midrule
	64 & 0.56\% & 27.97\% & \textbf{0.3608} & 1.7605\\
	128 & 1.25\% & 29.29\% & \textbf{0.4312} & 1.8740\\
	512 & 1.86\% & 32.37\% & \textbf{0.4769} & 2.0374\\
	1024 & 3.00\% & 35.92\% & \textbf{0.4892} & 2.0679\\
	2048 & 7.08\% & 39.19\% & \textbf{0.6052} & 2.1461\\
	\bottomrule
\end{tabular}
\vskip .2cm
\caption {ResNet-56 results on CIFAR-100 dataset with different batch sizes} 
\label{tab:title}	
\end {table}

\subsection{Main results and discussion}
Experimental results for VGG-16, ResNet-56 and LeNet-5 models on the different datasets based on various training batch sizes are given in Tables 1, 2, 3, 4, 5 \& 6. In these tables, the error rate and loss on the training set are reported as \textquoteleft train error\textquoteright~and \textquoteleft train loss\textquoteright; while the error rate and loss on the test set are given as \textquoteleft test error\textquoteright and \textquoteleft test loss\textquoteright. It is observed that for all models and datasets, the train errors, test errors, train loss and test loss all increase with increase in training batch size; train error and loss increase show optimization problem. \textit{Importantly, it is noted that though similar train errors can be obtained with an increase in training batch size, the \textquoteleft train loss\textquoteright~typically increases; see Tables 1, 2, 3, 4, 5 \& 6}.

\begin {table} [t!]	
\centering
\fontsize{9}{9}\selectfont
\begin{tabular}{|c|c|c|c|c|}
	\toprule
	Batch size & Train error & Test error & Train loss & Test loss \\	
	\midrule
	128 & 0.00\% & 7.74\% & \textbf{0.0733} & 0.3895\\
	512 & 0.00\% & 8.07\% & \textbf{0.0940} & 0.3940\\
	1024 & 0.00\% & 8.23\% & \textbf{0.1076} & 0.4011\\
	2048 & 0.23\% & 8.57\% & \textbf{0.1596} & 0.4142\\
	4096 & 1.48\% & 9.60\% & \textbf{0.2258} & 0.4395\\
	8192 & 5.81\% & 10.45\% & \textbf{0.3700} & 0.4786\\
	\bottomrule
\end{tabular}
\vskip .2cm
\caption {LeNet-5 results on Fashion-MNIST dataset with different batch sizes} 
\label{tab:title}	
\end {table}

\begin {table} [t!]	
\centering
\fontsize{9}{9}\selectfont
\begin{tabular}{|c|c|c|c|c|}
	\toprule
	Batch size & Train error & Test error & Train loss & Test loss \\	
	\midrule
	128 & 0.00\% & 0.44\% & \textbf{0.0156} & 0.0299\\
	512 & 0.00\% & 0.62\% & \textbf{0.0256} & 0.0418\\
	1024 & 0.00\% & 0.64\% & \textbf{0.0508} & 0.0701\\
	2048 & 0.01\% & 0.76\% & \textbf{0.1111} & 0.1316\\
	4096 & 0.16\% & 0.86\% & \textbf{0.1795} & 0.1961\\
	8192 & 0.66\% & 0.99\% & \textbf{0.2358} & 0.2428\\
	\bottomrule
\end{tabular}
\vskip .2cm
\caption {LeNet-5 results on MNIST dataset with different batch sizes} 
\label{tab:title}	
\end {table}

\begin{figure}[t!]
	\centering
	\includegraphics [width=0.65\textwidth]{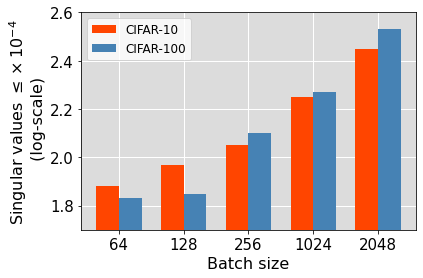}
	\caption {Near-rank loss of hidden representations with batch size for VGG-16, where $t_{th} = 10^{-4}$}
\end{figure}

\begin{figure}[t!]
	\centering
	\includegraphics [width=0.65\textwidth]{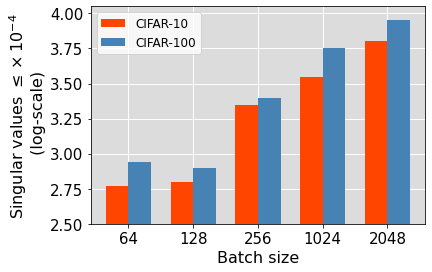}
	\caption {Near-rank loss of  hidden representations with batch size for ResNet-56, where $t_{th} = 10^{-4}$}
\end{figure}

\begin{figure}[t!]
	\centering
	\includegraphics [width=0.65\textwidth]{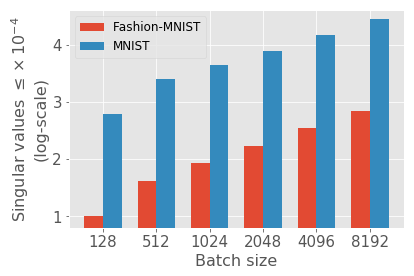}
	\caption {Near-rank loss of hidden representations with batch size for LeNet-5, where $t_{th} = 10^{-4}$}
\end{figure}

\begin{figure}[t!]
	\centering
	\includegraphics [width=0.65\textwidth]{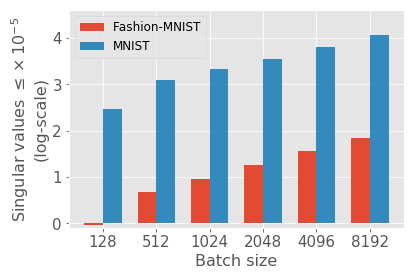}
	\caption {Near-rank loss of hidden representations with batch size for LeNet-5, where $t_{th} = 10^{-5}$}
\end{figure}

In Figures 1, 2 \& 3, the degree of near-rank loss of hidden units' activations for the different models with various batch sizes are shown as the number of singular values that are less or equal to the threshold value, $t_{th} = 10^{-4}$. Note that Figure 1 corresponds to results given in Tables 1 \& 2; Figure 2 corresponds to results given in Tables 3 \& 4; and Figure 3 corresponds to results given in Tables 5 \& 6. As such, it is seen that \textit{as the training batch size increases, so does the near-rank loss of hidden units' activations and training loss}. Further experimental results reveal that the generalization gap for models trained with large batches reported in Tables 1, 2, 3, 4, 5 \& 6 can be eliminated by simply switching to small batch sizes during training. That is, final test errors obtained are similar to those from models trained scratch using small batch sizes.

\subsection{Ablation studies}
\subsubsection{The degree of near-rank loss and with different singular value threshold}
Given the interesting results reported in Section 5.3 based on the cummulative near-rank loss of hidden representations with respect to different training batch sizes, we perform ablation studies to observe the impact of the chosen singular value threshold, $t_{th}$. For additional experiments, results for the singular value threshold of $t_{th} = 10^{-5}$ are reported in Figure 4. Using a smaller value of $t_{th}$ further emphasizes the degree of near-rank loss of the hidden representations in the models. For instance, compare Figure 3 and Figure 4. Note that ideally, we would like the singular value threshold to be zero, so that we can observe the exact rank loss of hidden representations. However, in practice, singular values may not be exactly zero, but extremely small so that near-rank loss that similarly plagues model training ensues. This scenario is similar to the concept of condition number that does not have to be infinite (i.e. the smallest singular value is zero), but be extremely large (i.e. the smallest singular value is very small) for the problem being solved to become chaotic. 

\subsubsection{The degree of near-rank loss and batch normalization}
The influence of batch normalization~\cite{16} on the degree of the near-rank loss of the hidden representations is studied. Table 7 and Table 8 report the results of the LeNet-5 models trained without batch normalization on the Fashion-MNIST and MNIST datasets using different batch sizes. It is seen from Table 7 and Table 8 that without batch normalization, the training loss of the models increases with batch size increase. This observation is similar to the models trained with batch normalization; compare Table 5 and Table 7, and Table 6 and Table 8. Finally, Figure 5 shows the degree of near-rank loss of the models given in Table 7 and Table 8 (i.e. trained without batch normalization). Again, it is seen that the degree of near-rank loss for the models overall increases with batch size increase.

\begin {table} [t!]	
\centering
\fontsize{9}{9}\selectfont
\begin{tabular}{|c|c|c|c|c|}
	\toprule
	Batch size & Train error & Test error & Train loss & Test loss \\	
	\midrule
	128 & 0.02\% & 8.91\% & \textbf{0.0785} & 0.4784\\
	512 & 1.16\% & 9.13\% & \textbf{1.1031} & 0.3651\\
	1024 & 3.27\% & 9.54\% & \textbf{0.1561} & 0.3371\\
	2048 & 5.36\% & 10.02\% & \textbf{0.2087} & 0.3385\\
	4096 & 7.76\% & 11.02\% & \textbf{0.2691} & 0.3560\\
	8192 & 9.87\% & 11.94\% & \textbf{0.3301} & 0.3873\\
	\bottomrule
\end{tabular}
\vskip .2cm
\caption {LeNet-5 without batch normalization results on Fashion-MNIST dataset with different batch sizes} 
\label{tab:title}	
\end {table}

\begin {table} [t!]	
\centering
\fontsize{9}{9}\selectfont
\begin{tabular}{|c|c|c|c|c|}
	\toprule
	Batch size & Train error & Test error & Train loss & Test loss \\	
	\midrule
	128 & 0.00\% & 0.68\% & \textbf{0.0190} & 0.0416\\
	512 & 0.00\% & 0.80\% & \textbf{0.0386} & 0.0646\\
	1024 & 0.02\% & 0.86\% & \textbf{0.0473} & 0.0741\\
	2048 & 0.14\% & 0.98\% & \textbf{0.0565} & 0.0786\\
	4096 & 0.56\% & 1.21\% & \textbf{0.0723} & 0.0889\\
	8192 & 1.47\% & 1.77\% & \textbf{0.0999} & 0.1057\\
	\bottomrule
\end{tabular}
\vskip .2cm
\caption {LeNet-5 without batch normalization results on MNIST dataset with different batch sizes} 
\label{tab:title}	
\end {table}

\begin{figure}[t!]
	\centering
	\includegraphics [width=0.65\textwidth]{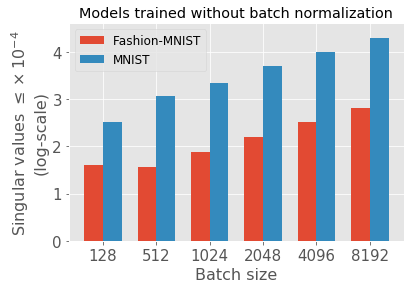}
	\caption {Near-rank loss of hidden representations with batch size for LeNet-5 trained without batch normalization, where $t_{th} = 10^{-4}$}
\end{figure}

\subsection{Results for linear DNN models}
In Section 5.3, results for non-linear DNNs (i.e. with non-linear activation functions) are discussed. In this Section, additional experiments are performed to show that similar results are obtained for linear DNNs (i.e. with linear activation functions). The results for linear LeNet-5 trained on Fashion-MNIST and MNIST datasets are given in Table 9 and Table 10, respectively. It is seen that as the batch size is increased, the achieved training losses increase, along with the test error rates. Furthermore, the number of singular values that equal or smaller than the threshold value $10^{-4}$ are show in Figure 6. Again, it is observed from Figure 6 that as the batch size is increased, the number of singular values greater than $10^{-4}$ consistently increases.

\begin {table} [t!]	
\centering
\fontsize{9}{9}\selectfont
\begin{tabular}{|c|c|c|c|c|}
	\toprule
	Batch size & Train error & Test error & Train loss & Test loss \\	
	\midrule
	128 & 4.46\% & 9.69\% & \textbf{0.1782} & 0.3467\\
	512 & 4.83\% & 10.26\% & \textbf{0.1946} & 0.3656\\
	1024 & 5.32\% & 10.36\% & \textbf{0.2428} & 0.3946\\
	2048 & 6.88\% & 10.51\% & \textbf{0.3594} & 0.4610\\
	4096 & 8.83\% & 10.78\% & \textbf{0.4994} & 0.5609\\
	8192 & 10.89\% & 12.08\% & \textbf{0.6187} & 0.6592\\
	\bottomrule
\end{tabular}
\vskip .2cm
\caption {Linear LeNet-5 results on Fashion-MNIST dataset with different batch sizes} 
\label{tab:title}	
\end {table}

\begin {table} [t!]	
\centering
\fontsize{9}{9}\selectfont
\begin{tabular}{|c|c|c|c|c|}
	\toprule
	Batch size & Train error & Test error & Train loss & Test loss \\	
	\midrule
	128 & 0.03\% & 1.03\% & \textbf{0.0269} & 0.0512\\
	512 & 0.04\% & 1.07\% & \textbf{0.0357} & 0.0644\\
	1024 & 0.10\% & 1.10\% & \textbf{0.0746} & 0.1008\\
	2048 & 0.40\% & 1.15\% & \textbf{0.1571} & 0.1758\\
	4096 & 1.01\% & 1.43\% & \textbf{0.2539} & 0.2606\\
	8192 & 1.72\% & 1.73\% & \textbf{0.3293} & 0.3270\\
	\bottomrule
\end{tabular}
\vskip .2cm
\caption {Linear LeNet-5 results on MNIST dataset with different batch sizes} 
\label{tab:title}	
\end {table}

\begin{figure}[t!]
	\centering
	\includegraphics [width=0.65\textwidth]{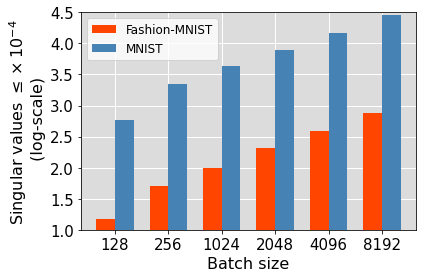}
	\caption {Near-rank loss of hidden representations with batch size for linear LeNet-5, where $t_{th} = 10^{-4}$}
\end{figure}

\section{The generalization gap problem in hindsight}
Following the results reported and discussed in Section 5, the different inferences that can be drawn for the generalization gap problem are presented in this section.
\subsection{Main insights into generalization gap}
\subsubsection{Training accuracy can be misleading}
Contrary to~\cite{16}, we find based on our investigation that the optimization evolutions are different for DNNs trained with different batch sizes. Thus, observing only the final training accuracies or error rates as in~\cite{16,19} can be misleading for studying generalization gap; rather, \textit{training loss is more appropriate for study. Models trained with small batch size generally reached smaller training losses than those trained with large batch sizes}; see Table 1 to Table 8. 

\subsubsection{Generalization is tied to optimization}  
DNNs trained with large batch sizes converge to higher training losses, suggesting that overfitting is not the problem. \textit{Rather, insufficient optimization that subsequently reflects in generalization gap is the problem}. Our argument aligns with the observation in the work~\cite{19} that DNNs trained with large batch sizes have parameters that are closer\footnote{Closer parameters after training corresponds to weaker optimization \label{2}} to the initialized values than models trained with small batch sizes.

\subsubsection{Generalization gap problem is not inherent} 
\textit{Generalization loss with batch size increase is not an inherent problem in the parameters space}, since we observe in this paper that switching from a large batch size to a small batch size during training allows optimization to proceed as if a small training batch size was used from the start of model training. Using Fashion-MNIST dataset, and taking the model trained with a batch size of 128 as the baseline for comparison, we observe the effect of switching from any arbitrary large batch size to 128 during training. The results are given in Table 11, where is seen that the final training losses are particularly similar such that we can conclude that there are no optimization problems. The testing errors are similar as well, which reflects that there is no generalization gap problem. Note that in Table 5, appreciable generalization loss is seen starting from a batch size of 4096. Our position that the optimization problem lies in the feature representation space, and not in the parameters space is corroborated by~\cite{20}, which observed no saddles points in the parameters space for large batch sizes. Hence, we argue that generalization gap is simply a result of an unsuitable selection of the batch size hyperparameter, just as convolution filters of small sizes (e.g. $3 \times 3$) are favoured for generalization performance~\cite{44} over filters of large sizes. 

\begin {table} [t!]	
\centering
\fontsize{9}{9}\selectfont
\begin{tabular}{|c|c|c|c|c|}
	\toprule
	Batch size & Train error & Test error & Train loss & Test loss \\	
	\midrule
	128 & 0.00\% & 7.74\% & \textbf{0.0733} & 0.3895\\
	4096 $\longrightarrow$ 128 & 0.00\% & 7.85\% & \textbf{0.0736} & 0.3908\\
	8192 $\longrightarrow$ 128 & 0.00\% & 7.50\% & \textbf{0.0736} & 0.3903\\
	16384 $\longrightarrow$ 128 & 0.00\% & 7.56\% & \textbf{0.0754} & 0.3835\\
	24576 $\longrightarrow$ 128 & 0.00\% & 7.64\% & \textbf{0.0742} & 0.3915\\
	\bottomrule
\end{tabular}
\vskip .2cm
\caption {LeNet-5 results on Fashion-MNIST dataset obtained by switching to a smaller batch size during training. Batch size of 128 is taken as baseline} 
\end {table}

\subsubsection{Increasing model width alleviates generalization gap} 
The work in~\cite{43} posits that increasing DNN width (i.e. $h^n_l$ for the hidden layer output, $\bm{H(x)}^l \in \mathbb{R}^{h^n_l \times b_s}$) alleviates the generalization gap problem. Interestingly, our analysis in Section 4.3 can be used to explain this phenomenon in that \textit{increasing $h^n_l$ causes the admissible distribution range for $\Ex \sigma_{min}(\bm{H(x)}^l)$ to increase as in~\ref{prop 2}}; that is, $P(\sigma_{min}(\bm{H(x)}^l)) > 10^{-4}$ increases. Subsequently, near-rank loss of $\bm{H(x)}^l$ is reduced, and optimization condition is improved so that the model generalizes better. An additional explanation is that increasing $h^n_l$ increases the number of parameters of the DNN so that the representational capacity of the model is increased, and thus may compensate for near-rank loss of hidden layer units and associated information loss. In contrast, increasing $b_s$ does not impact the number of model parameters, and thus, information loss due to the increased near-rank loss of hidden units (as observed in our analysis) cannot be compensated. 

\subsubsection{Generalization gap and near-rank loss of units' activations} 
\textit{A positive correlation is found between the near-rank loss of the hidden layer units' activations and batch size. i.e. Figure 1 to Figure 5}. The near-rank loss of hidden representations results in numerical instability~\cite{33}, and subsequently explains the optimization problem observed for large training batch sizes.


\subsection{Is the generalization gap problem really solved?}
The careful evaluation of methods that claim to have solved the generalization gap is advocated in this paper. Many training techniques such as ghost batch normalization~\cite{19} and RMSProp~\cite{42} in~\cite{41} are claimed to resolve the generalization gap of models trained with large batch size. Importantly, the drawback of these works~\cite{19,22,41} and others that claim to have solved the generalization gap problem is their failure to apply the same training techniques to the models trained with small batch sizes for fair comparison. Particularly, we argue that these training methods essentially improve model optimization, which subsequently reflects in better generalization. As such, models trained with small batch sizes can equally benefit from such training techniques. For example, the work~\cite{40} has already shown that ghost batch normalization improves the generalization performance of models trained with small batch sizes as well. Furthermore, the work in~\cite{39} has shown that the popular linear scaling of learning rates with batch size and gradual warm-up employed in~\cite{22} both improve the generalization performance of models trained with small batch size. Consequently, the claims in~\cite{19,22,41} and similar works are questionable and inconclusive, considering the simultaneous improvement for similar models trained using both small and large batche sizes.\\
Furthermore, we note an interesting recent work~\cite{56} with extensive experiments that contradicts the claim presented in~\cite{19} that training longer can resolve the generalization gap problem. 
In~\cite{56}, it was noted that using popular tricks such as learning warmup, learning rate scaling schemes, various optimizers and even exponentially longer training, the testing accuracy obtained using a small batch size could not be acheived when the batch size is increased beyond some certain limits. For MNIST, CIFAR-10 and Imagenet datasets, the batch size limits are 16000, 25000 and 128000, respectively~\cite{56}. Using a batch size of 819200 for Imagenet dataset on ResNet-50, and training for 10000 epochs, the best testing accuracy obtained was 71.8\%~\cite{56}. Interestingly, a testing accuracy of 76.1\% can be obtained using a batch size of 256 with less than 100 epochs as seen in the work~\cite{57}. Subsequently, all the evidence in this paper suggests that the generalization gap problem has not been indeed solved as claimed in the discussed papers. We particularly argue that a true solution would be one, where 
(i) applying similar training tricks does not improve the testing accuracy obtained using a small batch size, and (ii) using the full batch (all data samples) for training leads to no significant reduction in model tesing accuracy. 

\section{CONCLUSION}
\label{sec:conclusion} 
For the optimal utilization of computing resources and fast training, DNNs usually employ mini-batch gradient descent for training. However, an obstacle is the progressive generalization loss that ensues with training batch size increase. In this paper, we investigate from random matrix perspective, the source of the aforementioned  generalization loss. Our results reveal the correlation between the near-rank loss of hidden units' activations (that plagues optimization) and batch size increase. We posit that the generalization loss stems from an underlying difficulty of optimization that is associated with the near-rank loss of hidden units' activations tensors, which can be observed via singular value decomposition. Finally, our findings in conjunction with other recent works suggest that the claims of resolving the generalization gap problem by simply scaling the learning rate with batch size or applying ghost batch normalization are problematic and inconclusive. \\
Following the results of our analysis in the paper, an interesting direction for truly alleviating the generalization gap problem can involve tackling the increased near-rank loss of hidden representations with batch size increase. For instance, the addition of a calculated amount of noise to the different components of DNNs trained with large batch size. The randomness should generally reduce the degree of collinearity of hidden representations, and therefore reduce near-rank loss that is associated with generalization gap. Different ways of injecting various types of suitable noises in the hidden layers can be a promising and simpler approach for exploration.

\section*{Compliance with ethical standards}
\textbf{Conflict of interest}  The authors declare that they have no conﬂict of interest.

\section*{Data availability statement}
\textbf{Data availability}  All the datasets used in this paper are publicly available from third party sources, which are referenced including their weblinks.

\section*{Appendix}

\renewcommand{\thefigure}{A\arabic{figure}}
\setcounter{figure}{0}

\renewcommand{\thetable}{A\arabic{table}}
\setcounter{table}{0}

\renewcommand{\thesection}{A\arabic{section}}
\setcounter{section}{0}

\begin{figure*}[t!]
	\centering
	\begin{turn}{90}
		\begin{minipage}{7.3in}
			\centering
			\subfloat{{\includegraphics[width=3.5cm]{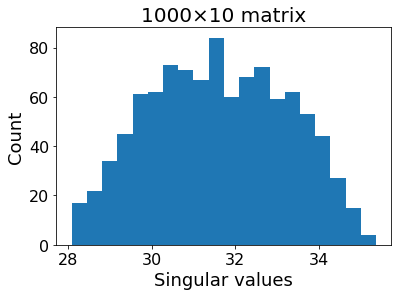} }}%
			\enspace
			\subfloat{{\includegraphics[width=3.5cm]{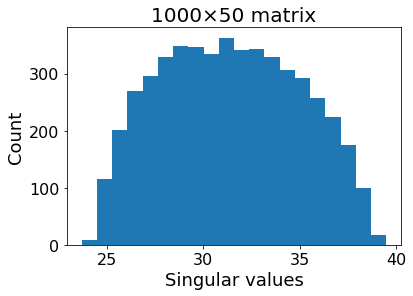} }}%
			\enspace
			\subfloat{{\includegraphics[width=3.5cm]{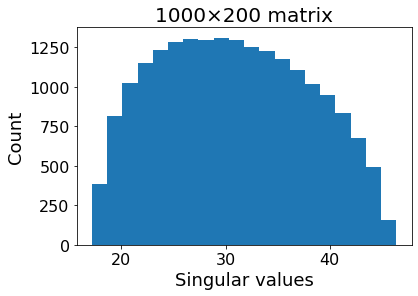} }}%
			\enspace
			\subfloat{{\includegraphics[width=3.5cm]{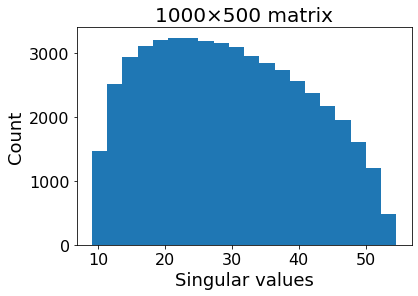} }}%
			\enspace
			\subfloat{{\includegraphics[width=3.5cm]{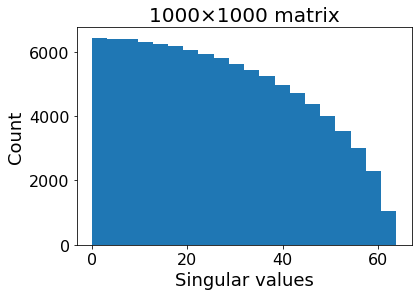} }} \\

			\subfloat{{\includegraphics[width=3.5cm]{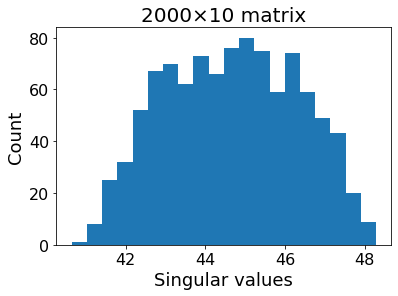} }}%
			\enspace
			\subfloat{{\includegraphics[width=3.5cm]{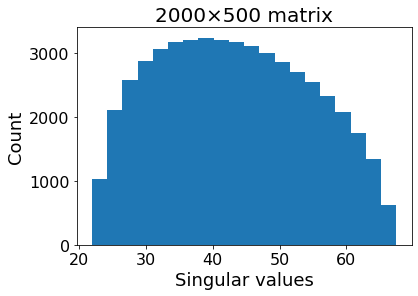} }}%
			\enspace
			\subfloat{{\includegraphics[width=3.5cm]{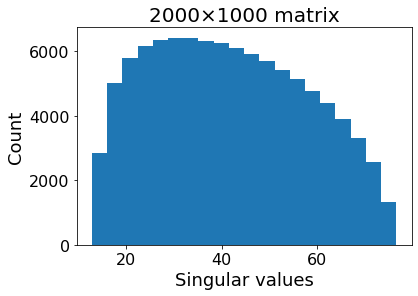} }}%
			\enspace
			\subfloat{{\includegraphics[width=3.5cm]{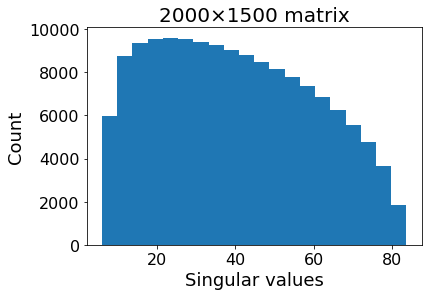} }}%
			\enspace
			\subfloat{{\includegraphics[width=3.5cm]{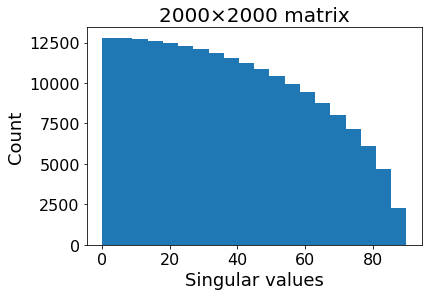} }} \\
			
			\subfloat{{\includegraphics[width=3.5cm]{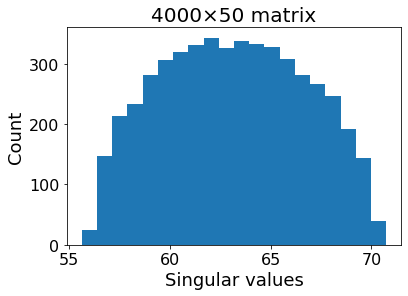} }}%
			\enspace
			\subfloat{{\includegraphics[width=3.5cm]{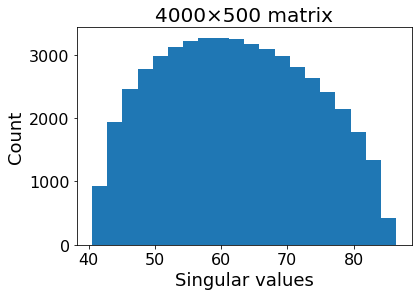} }}%
			\enspace
			\subfloat{{\includegraphics[width=3.5cm]{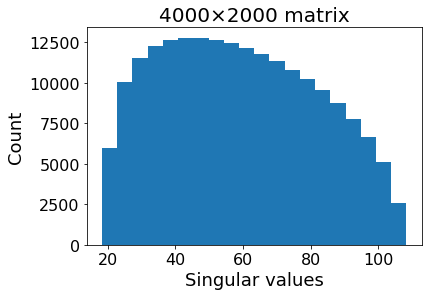} }}%
			\enspace
			\subfloat{{\includegraphics[width=3.5cm]{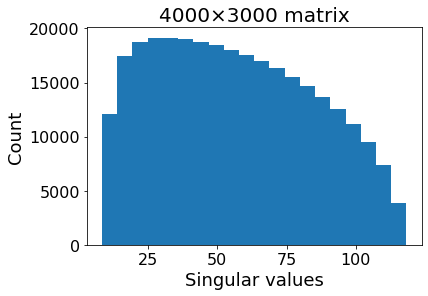} }}%
			\enspace
			\subfloat{{\includegraphics[width=3.5cm]{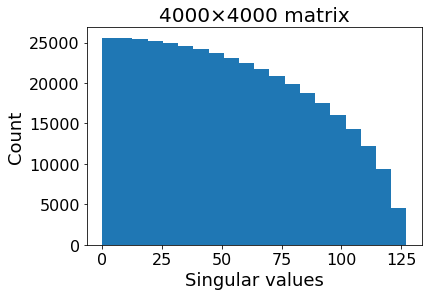} }}
			
			\caption {The singular values of random matrices with entries sampled from the standard normal distribution. One of the dimensions is fixed, while the other is varied. Each experiment is repeated 100 times. Top row: The maximum dimension of the matrcies is 1000. Middle                                                                                                                                                                                                                                                                                                                                                                                                                                                                                                                                                                                                                                                                                                                                                                                                                                                                                                                                                                                                                                                                                                                                                                                                                                                                                                                                                                                                                                                                                                                                                                                                                                                                                                                                                                                                                                                                                                                                                                                                                                                                                                                                                                                                                                                                                                                                                                                                                                                                                                                                                                                                                                                                                                                                                                                                                                                                     row: The maximum dimension of the matrcies is 2000. Bottom row: The maximum dimension of the matrcies is 4000.}
		\end{minipage}
	\end{turn}
\end{figure*}

\begin{figure*}[t!]
	\centering
	\begin{turn}{90}
		\begin{minipage}{7.3in}
			\centering
			\subfloat{{\includegraphics[width=3.5cm]{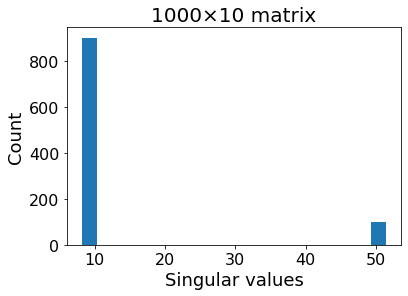} }}%
			\enspace
			\subfloat{{\includegraphics[width=3.5cm]{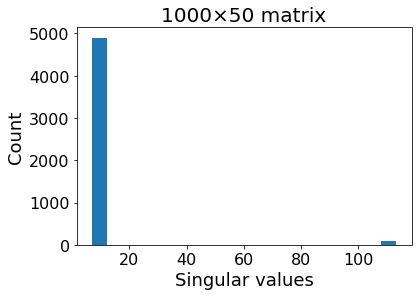} }}%
			\enspace
			\subfloat{{\includegraphics[width=3.5cm]{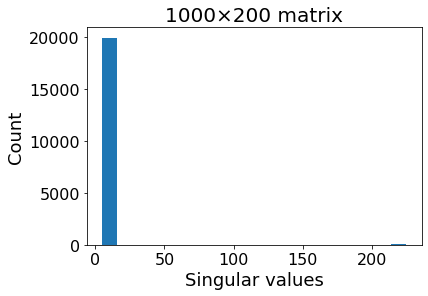} }}%
			\enspace
			\subfloat{{\includegraphics[width=3.5cm]{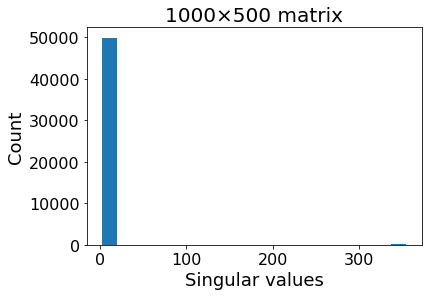} }}%
			\enspace
			\subfloat{{\includegraphics[width=3.5cm]{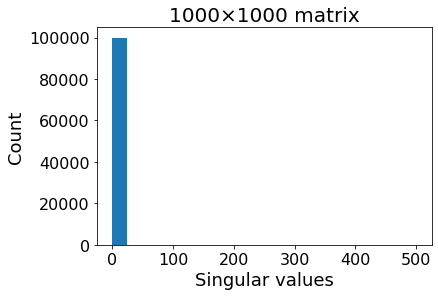} }} \\

			\subfloat{{\includegraphics[width=3.5cm]{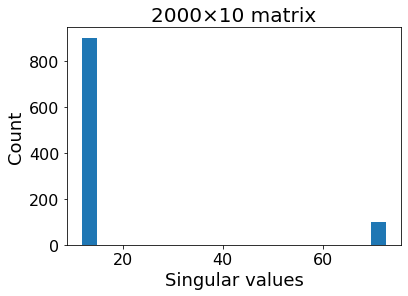} }}%
			\enspace
			\subfloat{{\includegraphics[width=3.5cm]{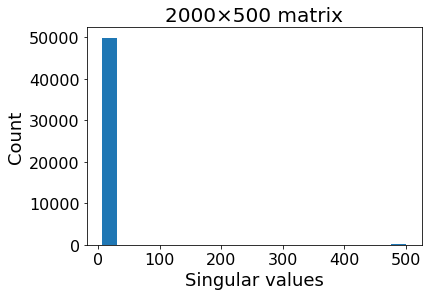} }}%
			\enspace
			\subfloat{{\includegraphics[width=3.5cm]{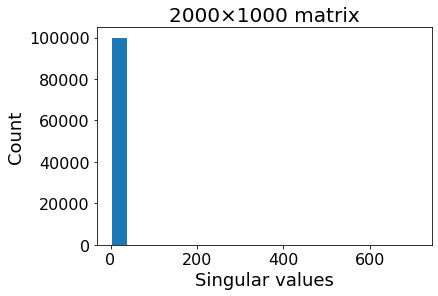} }}%
			\enspace
			\subfloat{{\includegraphics[width=3.5cm]{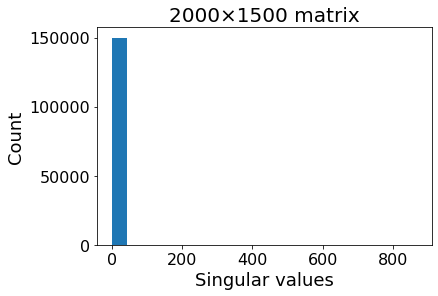} }}%
			\enspace
			\subfloat{{\includegraphics[width=3.5cm]{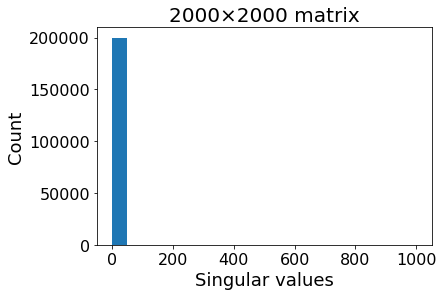} }} \\
			
			\subfloat{{\includegraphics[width=3.5cm]{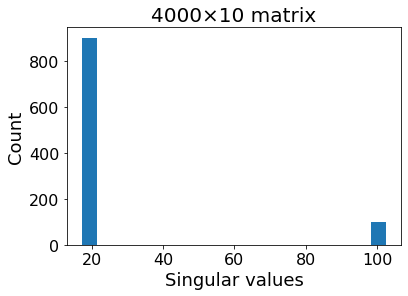} }}%
			\enspace
			\subfloat{{\includegraphics[width=3.5cm]{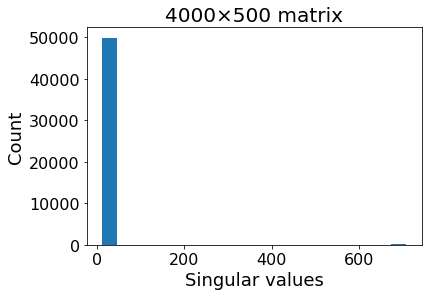} }}%
			\enspace
			\subfloat{{\includegraphics[width=3.5cm]{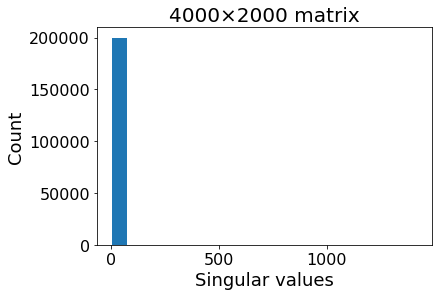} }}%
			\enspace
			\subfloat{{\includegraphics[width=3.5cm]{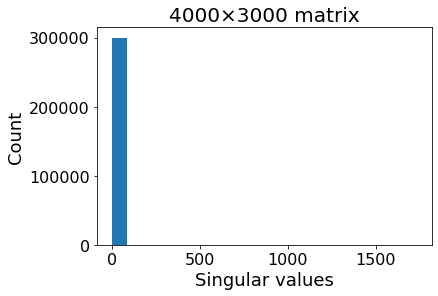} }}%
			\enspace
			\subfloat{{\includegraphics[width=3.5cm]{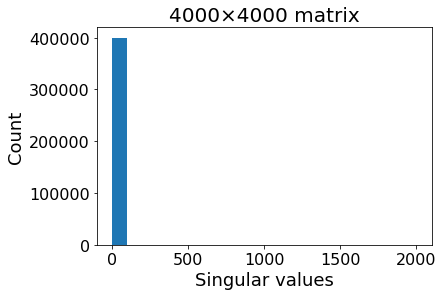} }}
			
			\caption {The singular values of random matrices with entries sampled from uniform distribution. One of the dimensions is fixed, while the other is varied. Each experiment is repeated 100 times. Top row: The maximum dimension of the matrcies is 1000. Middle row: The maximum dimension of the matrcies is 2000. Bottom row: The maximum dimension of the matrcies is 4000.}
		\end{minipage}
	\end{turn}
\end{figure*}

\begin{figure*}[t!]
	\centering
	\begin{turn}{90}
		\begin{minipage}{7.3in}
			\centering
			\subfloat{{\includegraphics[width=3.5cm]{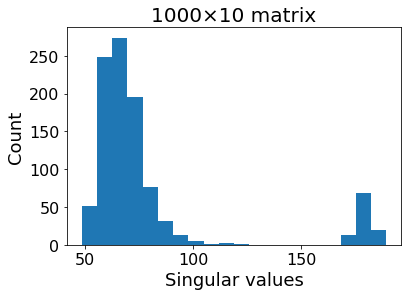} }}%
			\enspace
			\subfloat{{\includegraphics[width=3.5cm]{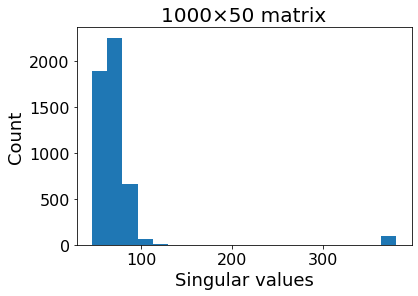} }}%
			\enspace
			\subfloat{{\includegraphics[width=3.5cm]{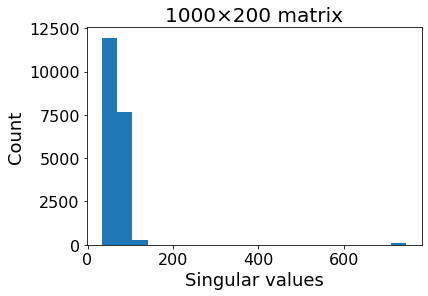} }}%
			\enspace
			\subfloat{{\includegraphics[width=3.5cm]{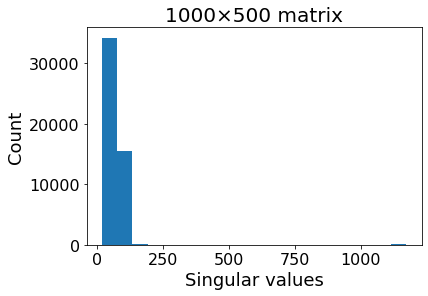} }}%
			\enspace
			\subfloat{{\includegraphics[width=3.5cm]{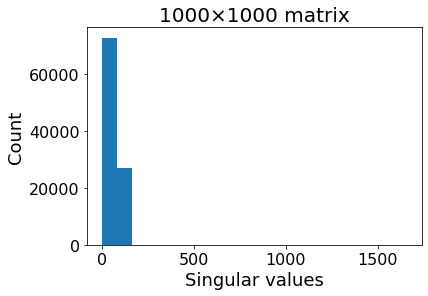} }} \\

			\subfloat{{\includegraphics[width=3.5cm]{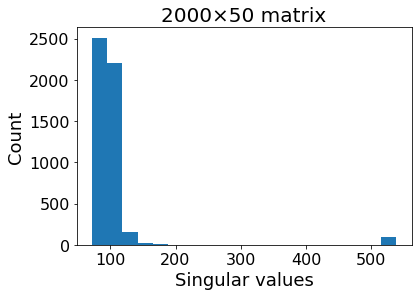} }}%
			\enspace
			\subfloat{{\includegraphics[width=3.5cm]{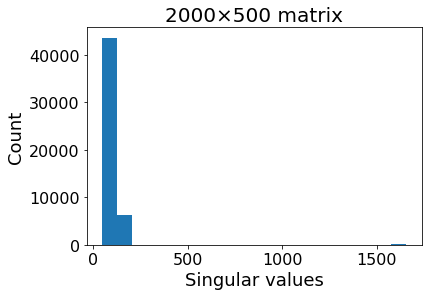} }}%
			\enspace
			\subfloat{{\includegraphics[width=3.5cm]{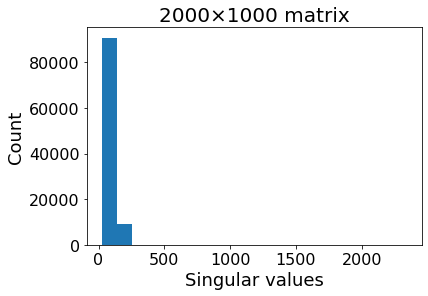} }}%
			\enspace
			\subfloat{{\includegraphics[width=3.5cm]{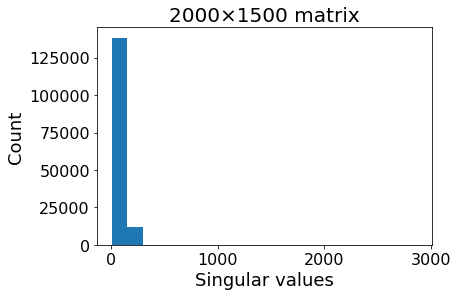} }}%
			\enspace
			\subfloat{{\includegraphics[width=3.5cm]{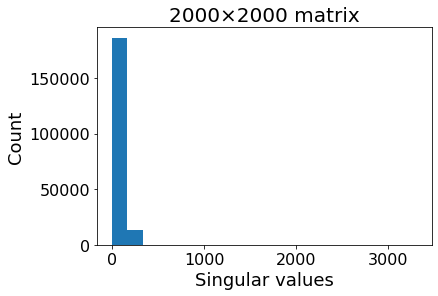} }} \\
			
			\subfloat{{\includegraphics[width=3.5cm]{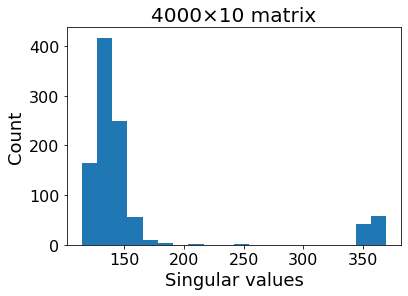} }}%
			\enspace
			\subfloat{{\includegraphics[width=3.5cm]{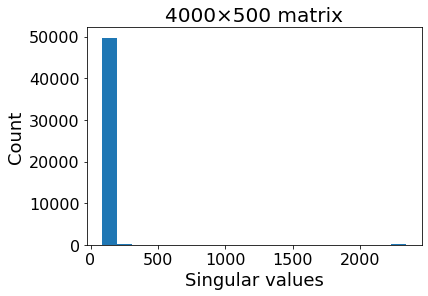} }}%
			\enspace
			\subfloat{{\includegraphics[width=3.5cm]{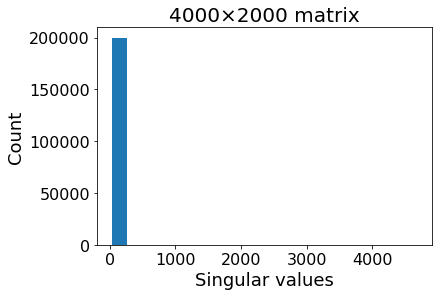} }}%
			\enspace
			\subfloat{{\includegraphics[width=3.5cm]{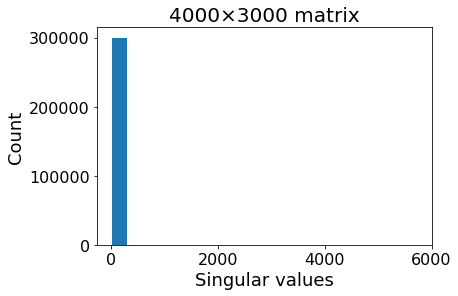} }}%
			\enspace
			\subfloat{{\includegraphics[width=3.5cm]{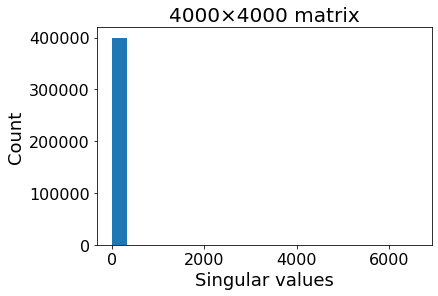} }}
			
			\caption {The singular values of random matrices with entries sampled from lognormal distribution of a mean value of zero and standard deviation of one. One of the dimensions is fixed, while the other is varied. Each experiment is repeated 100 times. Top row: The maximum dimension of the matrcies is 1000. Middle row: The maximum dimension of the matrcies is 2000. Bottom row: The maximum dimension of the matrcies is 4000.}
		\end{minipage}
	\end{turn}
\end{figure*}


\section{Singular values for random matrices}
\subsection{Distribution of the singular values of random matrices}
In this section, we empirically investigate the expected singular values for random matrices, which entries are drwan from Gaussian, uniform and lognormal distributions. \\
Namely, we show that given a random matrix $\bm{A} \in \mathbb{R}^{m \times n}: m \geq n$, many of the singular values of $\bm{A}$ becomes increasingly small  as $n$ is increased. Figure A1 shows the distributions of the singular values of random matrices with entries drawn from the standard normal distribution. For reinforcing the results, we experiment with $m=1000$, $m=2000$ and $m=4000$ for different values of $n$. Furthermore, each experiments are repeated 100 times. Given a specific value for $m$, it is seen that the number of small singular values increases as $n$ is increased. Similarly, Figure A2 shows the distributions of the singular values of random matrices with entries drawn from uniform distributions. Again, it is observed that for a given $m$ value, the many singular values become small as the value of $n$ is increased. For instance, for $m=1000$ and $n=10$ (i.e. $1000 \times 10$ matrix), some singular values are larger than 40. However, for $m=1000$ and $n=1000$ (i.e. $1000 \times 1000$ matrix), all the singular values are less than 40. Again, the distributions of the singular values of random matrices, which entries are sampled from lognormal distributions are shown in Figure A3. Finally, note that the singular values of $\bm{A}$ are the same as those of $\bm{A}^T$ (i.e. $\bm{A}$ transpose).

\begin {table} [t!]	
\centering
\fontsize{10}{10}\selectfont
\begin{tabular}{|c|c|}
	\toprule
	Matrix size & $\Ex \sigma_{min}(\bm{A})$ \\	
	\midrule
	$1000 \times 10$ & 29.9333 \\
	$1000 \times 50$ & 24.9032 \\
	$1000 \times 200$ & 17.6944 \\
	$1000 \times 500$ & 9.3842 \\
	$1000 \times 1000$ & 0.0214 \\
	\midrule
	$2000 \times 10$ & 42.0237 \\
	$2000 \times 500$ & 22.5139 \\
	$2000 \times 1000$ & 13.2249 \\
	$2000 \times 1500$ & 6.0893 \\
	$2000 \times 2000$ & 0.0142 \\
	\midrule
	$4000 \times 50$ & 60.5933 \\
	$4000 \times 500$ & 41.0717 \\
	$4000 \times 2000$ & 18.6311 \\
	$4000 \times 3000$ & 8.5605 \\
	$4000 \times 4000$ & 0.0103 \\
	\bottomrule
\end{tabular}
\vskip .2cm
\caption {Expected value of the minimum singular values for Gaussian random matrix. This corresponds to the matrices in Figure A1} 
\label{tab:title}	
\end {table}

\subsection{Expected minimum singular values for random matrices}
The expected minimum singular values of random matrices, which entries are drawn from Gaussian, uniform and lognormal distribution are investigated. First, we give the popular \textit{strong law of large numbers} that relates the expected value of random variables to the sample mean as follows
\begin{lemma}
	\namedProplabel{prop 1}{Proposition 1}
	(Strong law of large numbers): Let a sequence of $t$ independently and identically distributed random variables be $\{X_i \}_{i=1}^t$, the sample mean be $M_t$, and the expected value be $\Ex \bm{X}$. We have the following relation
	\begin{equation}
		P(\lim\limits_{t \to \infty} M_t = \Ex \bm{X}) = 1
	\end{equation}	
	
\end{lemma} 
\noindent \textit{Proof.} See~\cite{47,48} for derivations and detailed discussion. \myQED

Basically,~\ref{prop 1} shows that the sample mean approximates the expected value, as $t$ is increased. In this paper, we use $t=100$. Consequently, going forward, the sample mean of the minimum singular values of random matrices, which can be easily computed is used to approximate the expected value of minimum singular values of random matrices. \\
Again, given a random matrix $\bm{A} \in \mathbb{R}^{m \times n}: m \geq n$, using a fixed $m$, the expected value of the minimum singular values of $\bm{A}$, $\Ex \sigma_{min}(\bm{A})$, is studied. Specifically, we report the expected value of minimum singular values for the random matrices experimented with in Section A1.1. As such, the singular values distribution shown in Figure A1, Figure A2 and Figure A3 correspond to the expected values of the minimum singular values given in Table A1, Table A2 and Table A3, respectively. \\
It will be seen from Tables A1, A2 \& A3 that as the variable dimension (i.e. $n$) of the matrix increases, the expected value of minimum singular values decreases. This reflects that $\Ex \sigma_{min} (\bm{A}) \longrightarrow 0$, as $n$ becomes large.

\begin {table} [t!]	
\centering
\fontsize{10}{10}\selectfont
\begin{tabular}{|c|c|}
	\toprule
	Matrix size & $\Ex \sigma_{min}(\bm{A})$ \\	
	\midrule
	$1000 \times 10$ & 8.4324 \\
	$1000 \times 50$ & 7.2078 \\
	$1000 \times 200$ & 5.1265 \\
	$1000 \times 500$ & 2.7250 \\
	$1000 \times 1000$ & 0.0052 \\
	\midrule
	$2000 \times 10$ & 12.2200 \\
	$2000 \times 500$ & 6.5203 \\
	$2000 \times 1000$ & 3.8219 \\
	$2000 \times 1500$ & 1.7592 \\
	$2000 \times 2000$ & 0.0041 \\
	\midrule
	$4000 \times 50$ & 17.5597 \\
	$4000 \times 500$ & 11.8805 \\
	$4000 \times 2000$ & 5.3796 \\
	$4000 \times 3000$ & 2.4693 \\
	$4000 \times 4000$ & 0.0030 \\
	\bottomrule
\end{tabular}
\vskip .2cm
\caption {Expected value of the minimum singular values for uniform random matrix. This corresponds to the matrices in Figure A2} 
\label{tab:title}	
\end {table}

\begin {table} [t!]	
\centering
\fontsize{10}{10}\selectfont
\begin{tabular}{|c|c|}
	\toprule
	Matrix size & $\Ex \sigma_{min}(\bm{A})$ \\	
	\midrule
	$1000 \times 10$ & 56.1322 \\
	$1000 \times 50$ & 48.7673 \\
	$1000 \times 200$ & 36.0231 \\
	$1000 \times 500$ & 19.4127 \\
	$1000 \times 1000$ & 0.0422 \\
	\midrule
	$2000 \times 10$ & 84.1329 \\
	$2000 \times 500$ & 47.0961 \\
	$2000 \times 1000$ & 27.8701 \\
	$2000 \times 1500$ & 12.8825 \\
	$2000 \times 2000$ & 0.0307 \\
	\midrule
	$4000 \times 50$ & 17.5597 \\
	$4000 \times 500$ & 11.8805 \\
	$4000 \times 2000$ & 5.3796 \\
	$4000 \times 3000$ & 2.4693 \\
	$4000 \times 4000$ & 0.0227 \\
	\bottomrule
\end{tabular}
\vskip .2cm
\caption {Expected value of the minimum singular values for lognormal random matrix. This corresponds to the matrices in Figure A3} 
\label{tab:title}	
\end {table}

\bibliography{Oyedotun_Papadopoulos_Aouada_2022_REFS}

\begin{thebibliography}{10}
\providecommand{\url}[1]{#1}
\csname url@samestyle\endcsname
\providecommand{\newblock}{\relax}
\providecommand{\bibinfo}[2]{#2}
\providecommand{\BIBentrySTDinterwordspacing}{\spaceskip=0pt\relax}
\providecommand{\BIBentryALTinterwordstretchfactor}{4}
\providecommand{\BIBentryALTinterwordspacing}{\spaceskip=\fontdimen2\font plus
\BIBentryALTinterwordstretchfactor\fontdimen3\font minus
  \fontdimen4\font\relax}
\providecommand{\BIBforeignlanguage}[2]{{%
\expandafter\ifx\csname l@#1\endcsname\relax
\typeout{** WARNING: IEEEtran.bst: No hyphenation pattern has been}%
\typeout{** loaded for the language `#1'. Using the pattern for}%
\typeout{** the default language instead.}%
\else
\language=\csname l@#1\endcsname
\fi
#2}}
\providecommand{\BIBdecl}{\relax}
\BIBdecl

\bibitem{8}
Y.-R. Chien, J.-W. Chen, and S.~S.-D. Xu, ``A multilayer perceptron-based
  impulsive noise detector with application to power-line-based sensor
  networks,'' \emph{IEEE Access}, vol.~6, pp. 21\,778--21\,787, 2018.

\bibitem{9}
A.~A. Heidari, H.~Faris, I.~Aljarah, and S.~Mirjalili, ``An efficient hybrid
  multilayer perceptron neural network with grasshopper optimization,''
  \emph{Soft Computing}, vol.~23, no.~17, pp. 7941--7958, 2019.

\bibitem{7}
O.~Oyedotun and A.~Khashman, ``Prototype-incorporated emotional neural
  network.'' \emph{IEEE transactions on neural networks and learning systems},
  vol.~29, no.~8, pp. 3560--3572, 2018.

\bibitem{10}
W.~Schiffmann, M.~Joost, and R.~Werner, ``Comparison of optimized
  backpropagation algorithms.'' in \emph{ESANN}, vol.~93.\hskip 1em plus 0.5em
  minus 0.4em\relax Citeseer, 1993, pp. 97--104.

\bibitem{11}
B.~Zhou, A.~Lapedriza, A.~Khosla, A.~Oliva, and A.~Torralba, ``Places: A 10
  million image database for scene recognition,'' \emph{IEEE transactions on
  pattern analysis and machine intelligence}, vol.~40, no.~6, pp. 1452--1464,
  2017.

\bibitem{12}
A.~Krizhevsky, I.~Sutskever, and G.~E. Hinton, ``Imagenet classification with
  deep convolutional neural networks,'' in \emph{Advances in neural information
  processing systems}, 2012, pp. 1097--1105.

\bibitem{13}
A.~Bansal, A.~Nanduri, C.~D. Castillo, R.~Ranjan, and R.~Chellappa, ``Umdfaces:
  An annotated face dataset for training deep networks,'' in \emph{2017 IEEE
  International Joint Conference on Biometrics (IJCB)}.\hskip 1em plus 0.5em
  minus 0.4em\relax IEEE, 2017, pp. 464--473.

\bibitem{14}
K.~Osawa, Y.~Tsuji, Y.~Ueno, A.~Naruse, R.~Yokota, and S.~Matsuoka,
  ``Large-scale distributed second-order optimization using kronecker-factored
  approximate curvature for deep convolutional neural networks,'' in
  \emph{Proceedings of the IEEE Conference on Computer Vision and Pattern
  Recognition}, 2019, pp. 12\,359--12\,367.

\bibitem{15}
Y.~You, Z.~Zhang, C.-J. Hsieh, J.~Demmel, and K.~Keutzer, ``Imagenet training
  in minutes,'' in \emph{Proceedings of the 47th International Conference on
  Parallel Processing}.\hskip 1em plus 0.5em minus 0.4em\relax ACM, 2018, p.~1.

\bibitem{19}
E.~Hoffer, I.~Hubara, and D.~Soudry, ``Train longer, generalize better: closing
  the generalization gap in large batch training of neural networks,'' in
  \emph{Advances in Neural Information Processing Systems}, 2017, pp.
  1731--1741.

\bibitem{21}
S.~L. Smith, P.-J. Kindermans, C.~Ying, and Q.~V. Le, ``Don't decay the
  learning rate, increase the batch size,'' in \emph{International Conference
  on Learning Representations}, 2018, pp. 1--11.

\bibitem{22}
P.~Goyal, P.~Doll{\'a}r, R.~Girshick, P.~Noordhuis, L.~Wesolowski, A.~Kyrola,
  A.~Tulloch, Y.~Jia, and K.~He, ``Accurate, large minibatch sgd: Training
  imagenet in 1 hour,'' \emph{arXiv preprint arXiv:1706.02677}, 2017.

\bibitem{16}
N.~S. Keskar, D.~Mudigere, J.~Nocedal, M.~Smelyanskiy, and P.~T.~P. Tang, ``On
  large-batch training for deep learning: Generalization gap and sharp
  minima,'' in \emph{International Conference on Learning Representations},
  2017, pp. 1--16.

\bibitem{20}
Z.~Yao, A.~Gholami, Q.~Lei, K.~Keutzer, and M.~W. Mahoney, ``Hessian-based
  analysis of large batch training and robustness to adversaries,'' in
  \emph{Advances in Neural Information Processing Systems}, 2018, pp.
  4949--4959.

\bibitem{39}
D.~Masters and C.~Luschi, ``Revisiting small batch training for deep neural
  networks,'' \emph{arXiv preprint arXiv:1804.07612}, 2018.

\bibitem{17}
S.~Hochreiter and J.~Schmidhuber, ``Flat minima,'' \emph{Neural Computation},
  vol.~9, no.~1, pp. 1--42, 1997.

\bibitem{18}
{Hochreiter, Sepp and Schmidhuber, J{\"u}rgen}, ``Simplifying neural nets by
  discovering flat minima,'' in \emph{Advances in neural information processing
  systems}, 1995, pp. 529--536.

\bibitem{56}
Y.~You, Y.~Wang, H.~Zhang, Z.~Zhang, J.~Demmel, and C.-J. Hsieh, ``The limit of
  the batch size,'' \emph{arXiv preprint arXiv:2006.08517}, 2020.

\bibitem{58}
Y.~Wen, K.~Luk, M.~Gazeau, G.~Zhang, H.~Chan, and J.~Ba, ``An empirical study
  of large-batch stochastic gradient descent with structured covariance
  noise,'' in \emph{International Conference on Artificial Intelligence and
  Statistic}, 2020, pp. 1--15.

\bibitem{59}
J.~Wu, W.~Hu, H.~Xiong, J.~Huan, V.~Braverman, and Z.~Zhu, ``On the noisy
  gradient descent that generalizes as sgd,'' in \emph{International Conference
  on Machine Learning}.\hskip 1em plus 0.5em minus 0.4em\relax PMLR, 2020, pp.
  10\,367--10\,376.

\bibitem{60}
T.~Lin, L.~Kong, S.~Stich, and M.~Jaggi, ``Extrapolation for large-batch
  training in deep learning,'' in \emph{International Conference on Machine
  Learning}.\hskip 1em plus 0.5em minus 0.4em\relax PMLR, 2020, pp. 6094--6104.

\bibitem{61}
F.~He, T.~Liu, and D.~Tao, ``Control batch size and learning rate to generalize
  well: Theoretical and empirical evidence,'' in \emph{Advances in Neural
  Information Processing Systems}, vol.~32, 2019.

\bibitem{62}
L.~Ma, G.~Montague, J.~Ye, Z.~Yao, A.~Gholami, K.~Keutzer, and M.~Mahoney,
  ``Inefficiency of k-fac for large batch size training,'' in \emph{Proceedings
  of the AAAI Conference on Artificial Intelligence}, vol.~34, no.~04, 2020,
  pp. 5053--5060.

\bibitem{54}
A.~M. Saxe, J.~L. McClelland, and S.~Ganguli, ``Exact solutions to the
  nonlinear dynamics of learning in deep linear neural networks,'' in
  \emph{International Conference on Learning Representations}, 2014.

\bibitem{55}
K.~Kawaguchi, ``Deep learning without poor local minima,'' in \emph{Advances in
  Neural Information Processing Systems}, vol.~29, 2016, pp. 586--594.

\bibitem{49}
Y.~Zhou and Y.~Liang, ``Critical points of linear neural networks: Analytical
  forms and landscape properties,'' in \emph{International Conference on
  Learning Representations}, 2019.

\bibitem{50}
R.~Mulayoff and T.~Michaeli, ``Unique properties of flat minima in deep
  networks,'' in \emph{International Conference on Machine Learning}, 2020, pp.
  7108--7118.

\bibitem{52}
S.~Sonoda and N.~Murata, ``Transport analysis of infinitely deep neural
  network,'' \emph{The Journal of Machine Learning Research}, vol.~20, no.~1,
  pp. 31--82, 2019.

\bibitem{53}
T.~Laurent and J.~Brecht, ``Deep linear networks with arbitrary loss: All local
  minima are global,'' in \emph{International conference on machine learning},
  2018, pp. 2902--2907.

\bibitem{34}
M.~Rudelson and R.~Vershynin, ``Non-asymptotic theory of random matrices:
  extreme singular values,'' in \emph{Proceedings of the International Congress
  of Mathematicians 2010 (ICM 2010) (In 4 Volumes) Vol. I: Plenary Lectures and
  Ceremonies Vols. II--IV: Invited Lectures}.\hskip 1em plus 0.5em minus
  0.4em\relax World Scientific, 2010, pp. 1576--1602.

\bibitem{29}
G.~Bergqvist and E.~G. Larsson, ``The higher-order singular value
  decomposition: Theory and an application [lecture notes],'' \emph{IEEE Signal
  Processing Magazine}, vol.~27, no.~3, pp. 151--154, 2010.

\bibitem{45}
R.~Badeau and R.~Boyer, ``Fast multilinear singular value decomposition for
  structured tensors,'' \emph{SIAM Journal on Matrix Analysis and
  Applications}, vol.~30, no.~3, pp. 1008--1021, 2008.

\bibitem{25}
A.~Krizhevsky, V.~Nair, and G.~Hinton, ``Cifar-10, cifar-100 (canadian
  institute for advanced research),''
  \url{http://www.cs.toronto.edu/~kriz/cifar.html}, Last accessed, August 2020.

\bibitem{26}
Z.~research, ``Fashion-mnist handwritten dataset,''
  \url{https://github.com/zalandoresearch/fashion-mnist/}, Last accessed,
  August 2020.

\bibitem{27}
Y.~LeCun and C.~Cortes, ``Mnist handwritten digit database,''
  \url{http://yann.lecun.com/exdb/mnist/}, Last accessed, August 2020.

\bibitem{28}
X.~Ding, G.~Ding, J.~Han, and S.~Tang, ``Auto-balanced filter pruning for
  efficient convolutional neural networks,'' in \emph{Thirty-Second AAAI
  Conference on Artificial Intelligence}, 2018.

\bibitem{38}
C.~J. Shallue, J.~Lee, J.~Antognini, J.~Sohl-Dickstein, R.~Frostig, and G.~E.
  Dahl, ``Measuring the effects of data parallelism on neural network
  training,'' \emph{Journal of Machine Learning Research}, vol.~20, pp. 1--49,
  2019.

\bibitem{37}
S.~Ioffe and C.~Szegedy, ``Batch normalization: Accelerating deep network
  training by reducing internal covariate shift,'' in \emph{International
  Conference on Machine Learning}, 2015, pp. 448--456.

\bibitem{44}
K.~Simonyan and A.~Zisserman, ``Very deep convolutional networks for
  large-scale image recognition,'' in \emph{International Conference on
  Learning Representations}, 2015.

\bibitem{43}
L.~Chen, H.~Wang, J.~Zhao, D.~Papailiopoulos, and P.~Koutris, ``The effect of
  network width on the performance of large-batch training,'' in \emph{Advances
  in Neural Information Processing Systems}, 2018, pp. 9302--9309.

\bibitem{33}
F.~Cousseau, T.~Ozeki, and S.-i. Amari, ``Dynamics of learning in multilayer
  perceptrons near singularities,'' \emph{IEEE Transactions on Neural
  Networks}, vol.~19, no.~8, pp. 1313--1328, 2008.

\bibitem{42}
T.~Tieleman and G.~Hinton, ``Lecture 6.5-rmsprop: Divide the gradient by a
  running average of its recent magnitude,'' \emph{COURSERA: Neural networks
  for machine learning}, vol.~4, no.~2, pp. 26--31, 2012.

\bibitem{41}
T.~Akiba, S.~Suzuki, and K.~Fukuda, ``Extremely large minibatch sgd: Training
  resnet-50 on imagenet in 15 minutes,'' \emph{arXiv preprint
  arXiv:1711.04325}, 2017.

\bibitem{40}
C.~Summers and M.~J. Dinneen, ``Four things everyone should know to improve
  batch normalization,'' in \emph{International Conference for Learning
  Representations}, 2020, pp. 1--18.

\bibitem{57}
S.~Xie, R.~Girshick, P.~Doll{\'a}r, Z.~Tu, and K.~He, ``Aggregated residual
  transformations for deep neural networks,'' in \emph{Proceedings of the IEEE
  conference on computer vision and pattern recognition}, 2017, pp. 1492--1500.

\bibitem{47}
N.~Etemadi, ``An elementary proof of the strong law of large numbers,''
  \emph{Zeitschrift f{\"u}r Wahrscheinlichkeitstheorie und verwandte Gebiete},
  vol.~55, no.~1, pp. 119--122, 1981.

\bibitem{48}
I.~D. Dinov, N.~Christou, and R.~Gould, ``Law of large numbers: The theory,
  applications and technology-based education,'' \emph{Journal of Statistics
  Education}, vol.~17, no.~1, 2009.

\end{thebibliography}
\bibliographystyle{IEEEtran}

\end{document}